\documentclass[]{opendatalab}

\PassOptionsToPackage{numbers, compress}{natbib}

\usepackage[utf8]{inputenc} % allow utf-8 input
\usepackage[T1]{fontenc}    % use 8-bit T1 fonts
\usepackage{hyperref}       % hyperlinks
\usepackage{url}            % simple URL typesetting
\usepackage{booktabs}       % professional-quality tables
\usepackage{amsfonts}       % blackboard math symbols
\usepackage{nicefrac}       % compact symbols for 1/2, etc.
\usepackage{microtype}      % microtypography
\usepackage{xcolor}         % colors

\usepackage{multirow}
\usepackage{multicol}
\usepackage{colortbl}
\usepackage{adjustbox}
\usepackage{caption}
\usepackage{subcaption}
\usepackage{amsmath}
\usepackage{todonotes}
\usepackage{pifont}
\usepackage{array}
\usepackage{ragged2e}
\usepackage{booktabs}
\usepackage{wrapfig}

\usepackage{listings}
\usepackage[dvipsnames]{xcolor}
\definecolor{augProblemText}{HTML}{A6623A}
\definecolor{augPatternText}{HTML}{7A6B2C}
\usepackage[most]{tcolorbox}
\usepackage{tcolorbox}
\usepackage{enumitem}
\usepackage{graphicx}
\usepackage{arydshln}
\usepackage{subcaption}

\lstdefinestyle{customPython}{
  language=Python,
  basicstyle=\ttfamily\scriptsize,
  keywordstyle=\color{blue},
  numbers=none,
  stringstyle=\color{OliveGreen}
}

\tcbset{
  myframe/.style={
    width=\linewidth,
    boxrule=1pt,           % 边框粗
    colframe=black!75,     % 边框颜色
    colback=gray!10,       % 背景色
    arc=2mm,
    left=2mm, right=2mm,
    top=0.2mm, bottom=0.2mm,
    enhanced
  }
}

\newtcolorbox{infobox}[1][]{
  myframe,
  fonttitle=\bfseries,
  title=#1
}

\newtcblisting{codebox}{
  myframe,
  listing only,
  listing options={
    language=Python,
    basicstyle=\ttfamily\scriptsize,
    breaklines=true,
    numbers=none,
    numberstyle=\tiny\color{gray},
    frame=none         % ❗️关闭 listings 自带边框
  }
}

\definecolor{mygreen}{HTML}{097969}
\newcommand{\method}{{\texttt{Caco}}}
\setlist[itemize,enumerate,description]{nosep, left=0mm, itemsep=0.5mm}

\title{Scaling Code-Assisted Chain-of-Thoughts and Instructions for Model Reasoning}

\author[1,2\dag]{Honglin Lin}
\author[1\dag]{Qizhi Pei}
\author[1,2]{Xin Gao}
\author[1]{Zhuoshi Pan}
\author[1]{Yu Li}
\author[3]{Juntao Li}
\author[1]{Conghui He}
\author[1\ast]{Lijun Wu}

\affiliation[1]{OpenDataLab, Shanghai Artificial Intelligence Laboratory}
\affiliation[2]{Shanghai Jiao Tong University}
\affiliation[3]{Soochow University}

\abstract{
Reasoning capability is pivotal for Large Language Models (LLMs) to solve complex tasks, yet achieving reliable and scalable reasoning remains challenging. While Chain-of-Thought (CoT) prompting has become a mainstream approach, existing methods often suffer from uncontrolled generation, insufficient quality, and limited diversity in reasoning paths. 
Recent efforts leverage code to enhance CoT by grounding reasoning in executable steps, but such methods are typically constrained to predefined mathematical problems, hindering scalability and generalizability. 
In this work, we propose \texttt{Caco} (Code-Assisted Chain-of-ThOught), a novel framework that automates the synthesis of high-quality, verifiable, and diverse instruction-CoT reasoning data through code-driven augmentation. Unlike prior work, \texttt{Caco} first fine-tunes a code-based CoT generator on existing math and programming solutions in a unified code format, then scales the data generation to a large amount of diverse reasoning traces. Crucially, we introduce automated validation via code execution and rule-based filtering to ensure logical correctness and structural diversity, followed by reverse-engineering filtered outputs into natural language instructions and language CoTs to enrich task adaptability. This closed-loop process enables fully automated, scalable synthesis of reasoning data with guaranteed executability. 
Experiments on our created \texttt{Caco}-1.3M dataset demonstrate that \texttt{Caco}-trained models achieve strong competitive performance on mathematical reasoning benchmarks, outperforming existing strong baselines. Further analysis reveals that \texttt{Caco}’s code-anchored verification and instruction diversity contribute to superior generalization across unseen tasks. Our work establishes a paradigm for building self-sustaining, trustworthy reasoning systems without human intervention.
}

\date{\today}
\correspondence{Lijun Wu, \email{wulijun@pjlab.org.cn}}

\metadata[Code]{\url{https://github.com/LHL3341/Caco}}
\metadata[Equal contribution]{Honglin Lin, Qizhi Pei}

\begin{document}
\maketitle
\vspace{0.5cm}

\begin{figure*}[h]
\vspace{-0.6cm}
    \centering
    \begin{subfigure}{0.45\linewidth}
        \includegraphics[width=\linewidth]{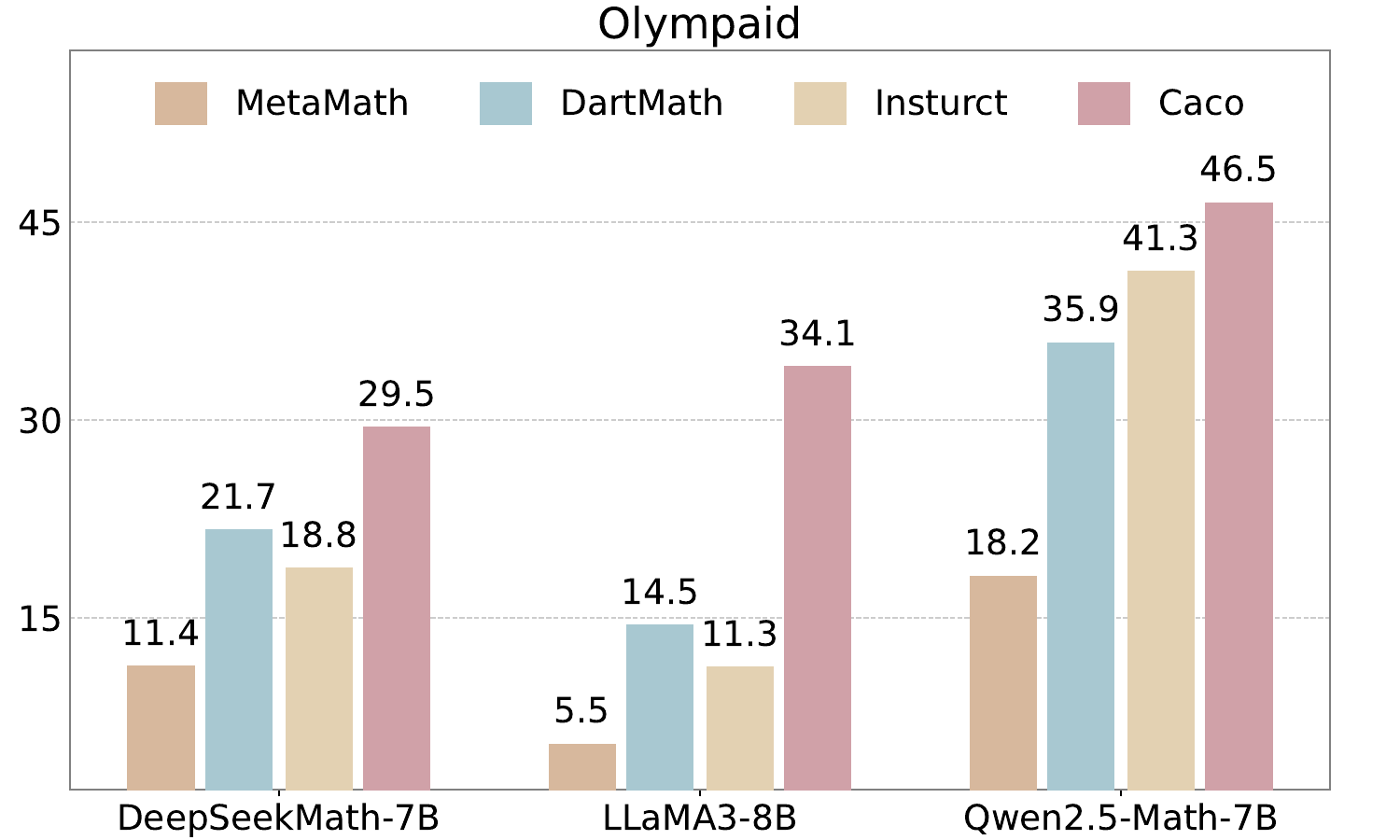}
    \end{subfigure}
    \begin{subfigure}{0.45\linewidth}
        \includegraphics[width=\linewidth]{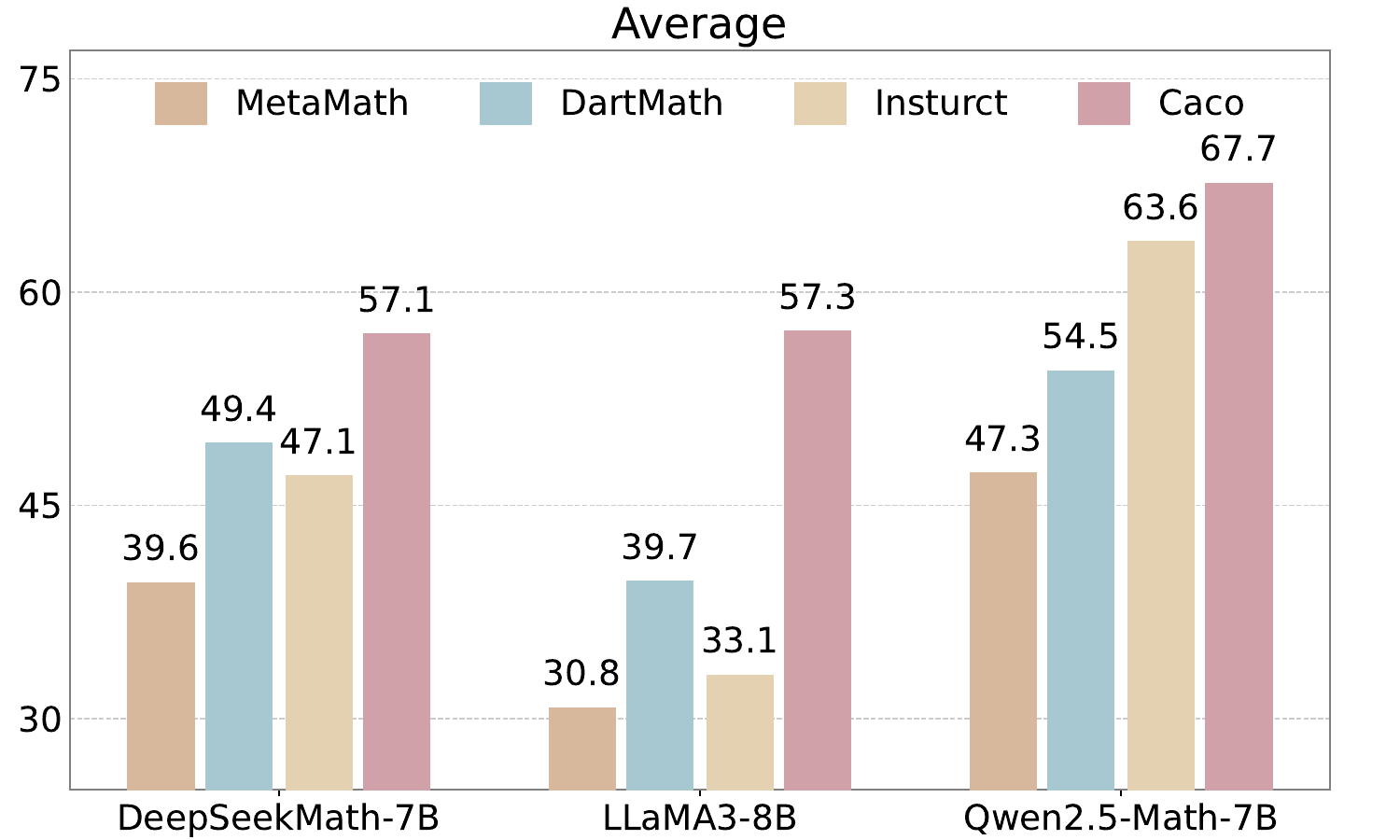}
    \end{subfigure}
    \label{fig:sota}
    \caption{Overview of \method{} results. \method{} shows superior performance on Olympiad Bench and on average than baseline methods.}
\vspace{-0.3cm}
\end{figure*}

\vspace{-1em}
\section{Introduction}
\label{sec:intro}
The advent of Large Language Models (LLMs)~\cite{qwen25_math,deepseekmath,mathstral} has revolutionized domains requiring complex reasoning, such as mathematics, code, and algorithmic problem-solving~\cite{cot_reasoning,llm_reasoning_survey,llm_math_survey}. Recent LLMs demonstrate remarkable capabilities in generating step-by-step solutions through Chain-of-Thought (CoT)~\cite{cot_reasoning,wizardmath,metamath} prompting, where intermediate reasoning steps are explicitly articulated before final answers. This paradigm has become instrumental in tasks like mathematical problem solving and program synthesis, where systematic logic decomposition is critical. 
A prevalent strategy involves generating long CoT sequences~\cite{openai_o1,deepseek_r1,qwq-32b-preview,qwen3} to mimic human-like deliberation. 

However, these CoT approaches predominantly rely on natural language reasoning traces, which suffer from several limitations. (1) \textit{Unverifiability}, since natural language reasoning is not executable, errors in intermediate steps may propagate and lead to incorrect conclusions; 
(2) \textit{Scalability constraints}, high-quality CoT data typically requires manual annotation, making it difficult to scale to diverse problem domains.

To address these issues, recent works have explored code-assisted reasoning~\cite{rstar_math,mathcoder,mathgenie,tora}, where reasoning steps are grounded in executable code snippets (e.g., Python codes or algorithm sketches). By translating natural language logic into formal code, these methods enable automatic verification through code execution. Preliminary studies~\cite{rstar_math} demonstrate that code-verified CoT can reduce hallucination and improve answer accuracy. 
However, existing implementations struggle to generalize beyond predefined mathematical problems, limiting their adaptability and scalability~\cite{openmathinstruct,mathcoder,tora}.

In this work, we introduce \texttt{Caco}, a scalable code-assisted CoT and instruction generation framework designed to automate the production of high-quality reasoning training data through code-anchored refinement. A core innovation of \texttt{Caco} lies in its fine-tuning of a base LLM on a compact set of structured code CoT demonstrations, enabling the model to learn systematic code reasoning solutions. Leveraging this fine-tuned LLM, we generate large-scale candidate code-based CoT solutions, which are subsequently refined via an automated verification engine. This engine executes code snippets, verifies logical consistency, and enforces diversity in reasoning patterns. Finally, the validated code solutions are translated back into natural language instructions and the corresponding language CoTs, yielding instruction-aligned data pairs that establish bidirectional alignment between code and textual reasoning paths.
The \texttt{Caco} generated natural language CoT offers several advantages.
(1) \textit{Scalability}: Through these model-generated synthetic code CoTs, we eliminate reliance on manual annotation of the aligned language CoTs, enabling the creation of millions of high-quality reasoning traces (e.g., our \texttt{Caco}-1.3M dataset); 
(2) \textit{Verifiability}: Not only are the answers guaranteed to be correct for the augmented instructions, but the executable and automatic validation of intermediate steps of Code CoTs also ensures the aligned language CoTs to be correct solutions. 
(3) \textit{Diversity}: By harnessing the fine-tuned LLM's generative capacity and sampling mechanism, \texttt{Caco} produces varied reasoning paths as well as the instructions, enhancing generalization across different problem types.

We evaluate \texttt{Caco} through extensive experiments on standard mathematical reasoning benchmarks. Models fine-tuned using our \texttt{Caco}-1.3M dataset achieve strong competitive performance; for example, attaining 92.6\% accuracy on GSM8K and 82.4\% on MATH, significantly outperforming prior approaches. \texttt{Caco} also exhibits strong generalization, the trained model maintains 67.7\% accuracy on average over multiple benchmarks, surpassing comparable methods by a margin exceeding 7.9\%. Further analysis confirms that \texttt{Caco}-generated CoT data preserves high diversity and scalability.
Beyond advancing superior performance in mathematical reasoning, our work establishes a generalizable framework for developing self-improving and verifiable LLMs across algorithmic domains.

\section{Related Work}
\subsection{Data Augmentation for Mathematical Reasoning}
A wide range of recent efforts have explored different strategies for constructing instruction-tuning datasets tailored to mathematical reasoning~\cite{mathcoder,lin2025metaladder,yue2024mammoth}. For example, WizardMath~\cite{wizardmath}, MetaMath~\cite{metamath}, Orca-Math~\cite{mitra2024orcamath}, MMIQC~\cite{liu2024mmiqc}, and MathFusion~\cite{pei2025mathfusion} enhance answers and rationales for seed problems through prompt engineering and reinforcement learning techniques.
KPMath~\cite{kpmath}, MathScale~\cite{mathscale}, and ScaleQuest~\cite{scale_quest} generate new problems from scratch by extracting mathematical concepts and topical structures.
MAmooTH2~\cite{yue2024mammoth2} and Numina-Math~\cite{numinamath} construct instruction-tuning datasets by collecting and curating large-scale data from the web.
DART-Math~\cite{dartmath} applies rejection sampling based on problem difficulty to ensure the quality of generated solutions.
\method{} also falls within this scope and uses code as a scalable medium to generate diverse mathematical problems.

\subsection{Code Integration for Enhanced Reasoning}
LLMs often make calculation errors in complex mathematical reasoning (e.g., computing eigenvalues) when using CoT prompting~\cite{pot,pal}. To address this, methods such as Program of Thoughts (PoT)~\cite{pot}, Program-Aided Language models (PAL)~\cite{pal}, and Code-based Self-Verification (CSV)~\cite{csv} are proposed to prompt LLMs to generate executable code, leveraging external code interpreters for accurate computation. 
As open-source models improve, code-integrated data for post-training has gained attention. OpenMathInstruct-1~\cite{openmathinstruct}, TORA~\cite{tora}, MathCoder~\cite{mathcoder,mathcoder2}, MegaMath~\cite{megamath}, and DotaMath~\cite{dotamath} embed code within natural language, enabling more robust reasoning.
MAmmoTH~\cite{yue2024mammoth} introduces MathInstruct, a hybrid of CoT and PoT datasets, allowing for different reasoning strategies for different problems.
rStar-Math~\cite{rstar_math} generates paired natural language rationales and Python code, keeping only verified executable steps. 
CodeI/O~\cite{codeio} distills diverse reasoning patterns embedded in code by transforming it into a code input-output prediction format.
MathGenie~\cite{mathgenie} synthesizes math problems and code-integrated solutions through solution augmentation, question back-translation, and verification-based filtering.
Unlike previous works, our \method{} leverages a code generation model to ensure both \textit{scalability} and \textit{verifiability}, while additionally introducing algorithmic problem types to promote greater \textit{diversity} in problem coverage.

\section{Method}
\begin{figure*}[t]
    \centering
    \vspace{-0.5cm}
    \includegraphics[width=0.9\linewidth]{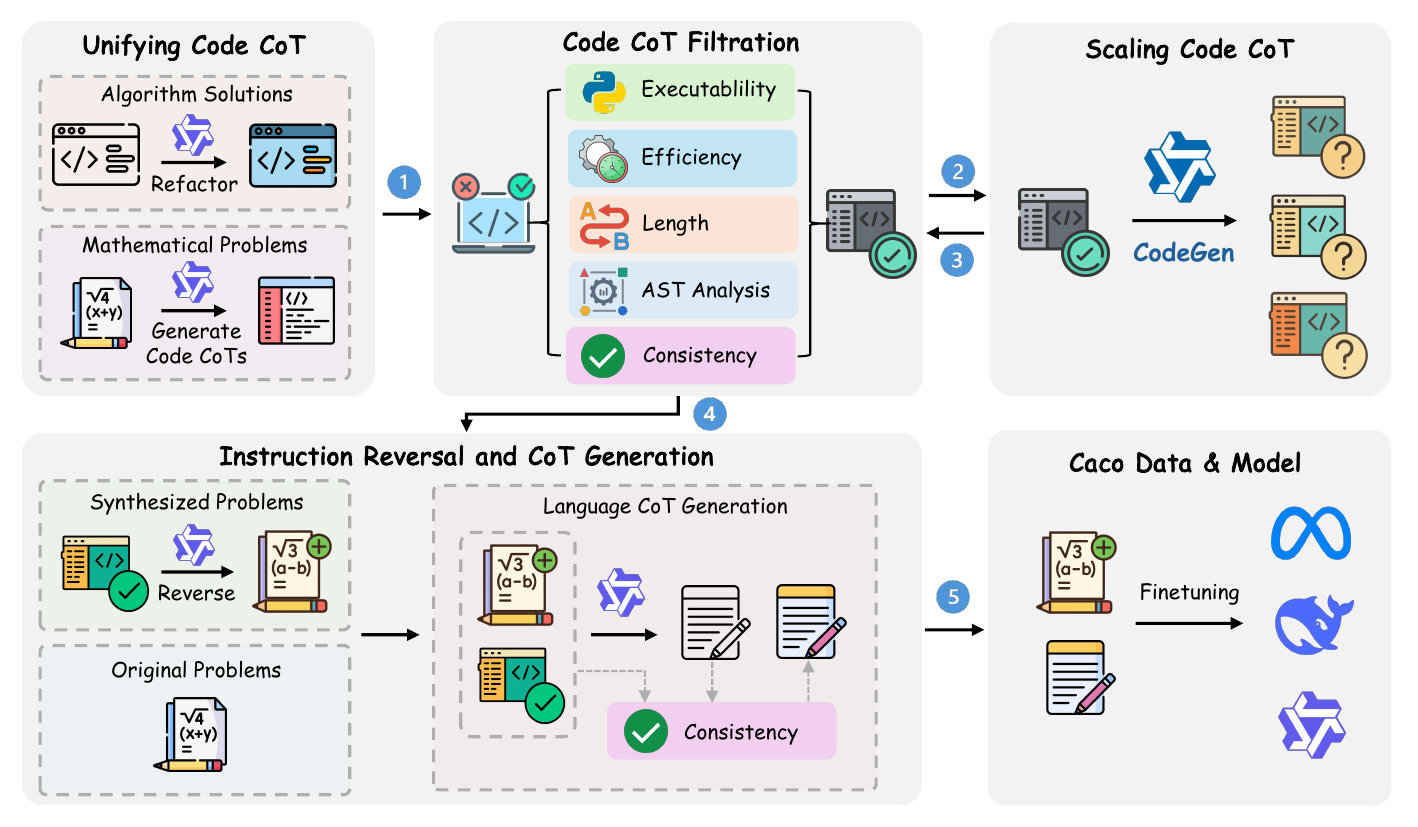}
    \caption{An overview framework of \texttt{Caco} data generation, including unifying Code CoT, scaling Code CoT with \texttt{CodeGen}, and instruction reversal and language CoT generation.}
    \label{fig:overview}
    \vspace{-0.5cm}
\end{figure*}

\paragraph{Overview.} Figure~\ref{fig:overview} presents the overall framework of \method. We begin by abstracting each problem's solution into an executable code template. Based on this, we fine-tune a problem generation model (\texttt{CodeGen}) to learn diverse reasoning strategies by extending these templates. Sampling from the trained model yields a large number of new programs, each representing a unique solution pattern for a particular family of problems. 
Each code pattern is back-translated into concrete mathematical problems and corresponding step-by-step solutions.
Only the instances where the natural language answer matches the code output are retained.
Details regarding the prompt formulations, model specifications, and the criteria for filtering Code CoTs are provided in Appendix~\ref{apx:datagen}.

\subsection{Unifying Code CoT}

To improve the quality, consistency, and verifiability of CoT reasoning for math problems, we explore a unified Code CoT representation. Motivated by prior findings that code data can enhance mathematical reasoning in language models~\cite{chen2022program,yue2024mammoth}, we collect and standardize Code CoTs from both mathematical and algorithmic domains. Specially, we use a general LLM $G_{p\to c}:\mathcal{P}\to\mathcal{C}$ to map each problem $p$ to code $c$ and an executor $F$ that returns the correct answer $a^*$ upon running $c$. We retain only verified traces; namely, the seed set is $\mathcal{C}_{\text{seed}}=\{\,G_{p \to c}(p)\mid F(c)=a^*\,\}$. This unified representation not only improves interpretability and execution fidelity but also lays the groundwork for scalable data generation and model training.

\noindent{\textbf{Mathematical Problems.}}
We collected a broad set of mathematical problems from multiple sources to ensure diversity, such as the MATH dataset~\cite{hendrycks2021math} (7.5K), DeepScaleR~\cite{deepscaler} (40K), and BigMath~\cite{bigmath} (251K). These problems vary in complexity and format; some are accompanied by natural language CoT explanations, while others are not. 
To unify their representation, we convert each solution into a structured Python program following a generic template (See Prompt~\ref{pmt:math2code}).
This template encodes problem inputs as dictionaries and defines problem-solving logic through explicit function calls. It supports a wide range of reasoning types—arithmetic, algebraic, geometric, probabilistic—while enabling direct execution for correctness verification. 

For example, consider the problem:

\begin{tcolorbox}
\vspace{-0.15cm}
\textit{George has an unfair six-sided die. The probability that it rolls a 6 is $\frac{1}{2}$, and the probability that it rolls any other number is $\frac{1}{10}$. What is the expected value of the number shown when this die is rolled?}
\vspace{-0.15cm}
\end{tcolorbox}

We transform its solution into the following code representation:

\vspace{-1.5mm}
\begin{codebox}
def expected_value(probabilities, values):
    return sum(p * v for p, v in zip(probabilities, values))

probabilities = [1/10, 1/10, 1/10, 1/10, 1/10, 1/2] # Probabilities for 1, 2, 3, 4, 5, 6
values = [1, 2, 3, 4, 5, 6] # Values on the die

input = {"probabilities": probabilities, "values": values}
output = expected_value(**input)
print(output)
\end{codebox}
\vspace{-1.5mm}

This standardized representation ensures structural consistency across different problem types and facilitates easier interpretation by both models and humans.

\noindent{\textbf{Algorithmic Problems.}}
In parallel, we incorporate algorithmic problems as an additional source of structured reasoning. We sample 40K problems from the Kodcode~\cite{kodcode} dataset, covering key algorithmic domains such as sorting, searching, and dynamic programming. These problems typically come with code-level solutions and brief natural language comments, providing a native form of Code CoT.
To ensure consistency across data sources, we normalize all algorithmic solutions into the same Python-based template used for mathematical problems. This standardization enables joint training and evaluation under a single format. The conversion prompts are described in Prompt~\ref{pmt:unify_code}.

\noindent{\textbf{Unified Seed Code CoTs.}}
\label{sec:unify_cot}
After Code CoT generation, we perform rigorous post-processing to ensure quality. Following the procedure described in Section~\ref{method:filter}, we validate each code sample through execution: only programs that run successfully, produce correct outputs, and conform to the standardized format are retained. This filtering yields a curated seed corpus of 146K high-quality Code CoT instances (122K Math + 24K Code). Among these, 109K problems originally had solutions, which we refer to as Seed109K in the experiments.
The resulting dataset provides a robust foundation for training models to enable effective generation of verifiable and scalable CoT reasoning in executable form in Section~\ref{sec:codegen}.

\begin{figure*}[t]
    \centering
    \vspace{-0.5cm}
    \includegraphics[width=0.9\linewidth]{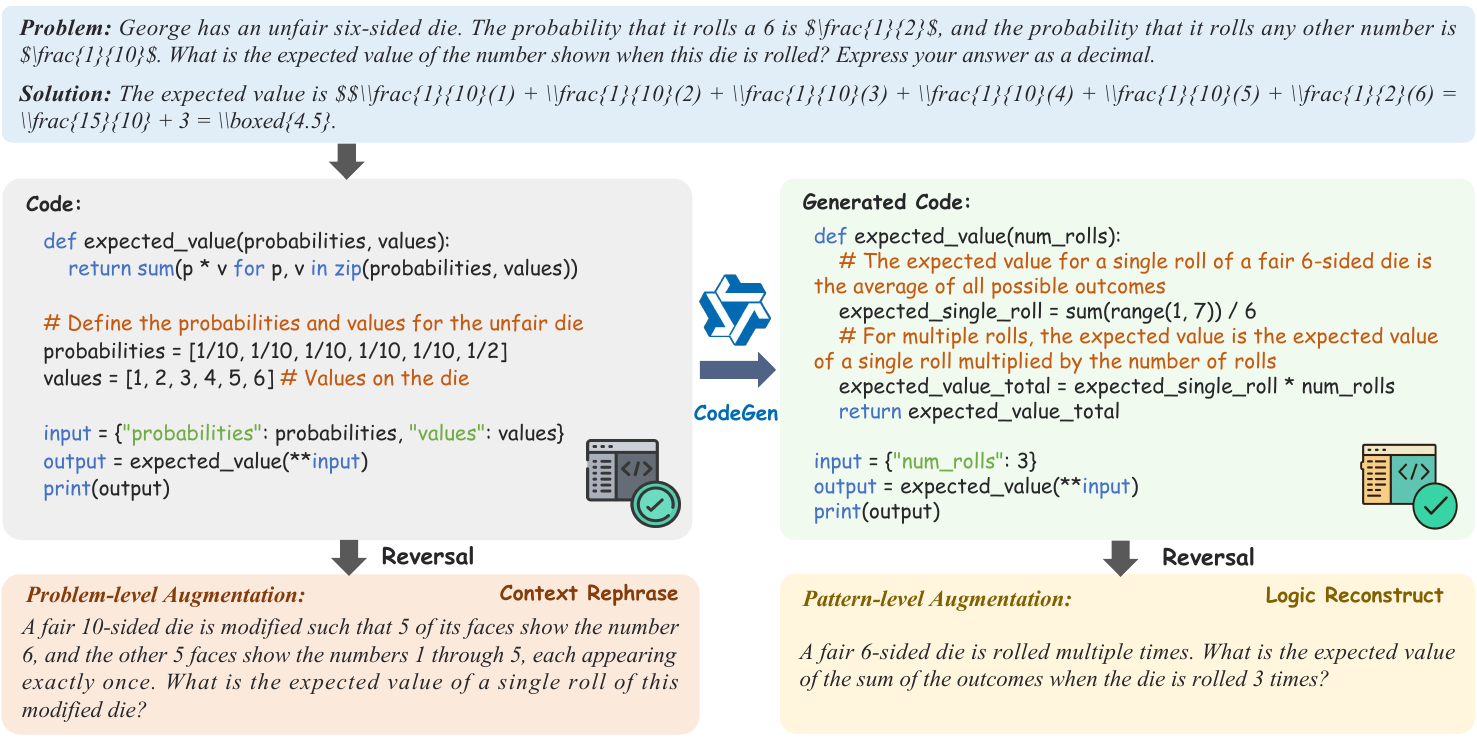}
    \caption{A case of one problem with its Code CoT. We demonstrate two augmentations, where problem-level augmentation refers to the original Code CoT can be back-translated into multiple question variants, and pattern-level augmentation means our \texttt{CodeGen} is capable of generating novel Code CoTs that generalize beyond the original seed patterns.}
    \label{fig:case}
    \vspace{-0.5cm}
\end{figure*}

\subsection{Scaling Code CoT with CodeGen}
\label{sec:codegen}

To scale the generation of high-quality code-based reasoning chains, we leverage the seed Code CoT dataset introduced previously to train a dedicated Code CoT generation model, \texttt{CodeGen}, so as to enable automated synthesis of executable, diverse, and logically coherent Code CoTs at scale. By training a model to internalize the structure and logic of our unified format, we facilitate the creation of new reasoning traces without relying on costly human annotations or handcrafted solutions.

\noindent{\textbf{Training CodeGen on Unified Code CoTs.}}
We fine-tune a unconditional \texttt{CodeGen} $U_\theta$ on $\mathcal{C}_{\text{seed}}$ to model the distribution of valid reasoning programs. 
\begin{equation}
\label{eq:uncond-mle}
\min_{\theta}\;\mathcal{L}(\theta)\;=\;\sum_{c\in\mathcal{C}_{\text{seed}}}\sum_{t=1}^{|c|}\log p_\theta\!\bigl(c_t\,\big|\,c_{<t}\bigr).
\end{equation}
The resulting model, \texttt{CodeGen}, is designed to generalize the reasoning patterns embedded in our dataset and produce structurally consistent code-based CoTs for both mathematical and algorithmic problems. 
Fine-tuning is conducted using a pretty simple prompt described in Appendix Table~\ref{pmt:codegen_train}. 
Notably, training uses only Code CoTs, without problem contexts or requirements, focusing on internalizing the reasoning trace space rather than specific problem-solution pairs.
In this way, we aim to largely explore the diverse Code CoTs in the generation phase.

\noindent{\textbf{Large-Scale CoT Generation via Sampling.}}
After fine-tuning, we employ \texttt{CodeGen} to generate a large number of new Code CoT samples $\mathcal{C}_{\text{samp}}=\{c' \sim U_\theta$\}. Using temperature sampling, we generate multiple candidate programs use prompt in Appendix Table~\ref{pmt:codegen_sample} (same as training \texttt{CodeGen}). This sampling-based approach introduces stochasticity into the decoding process, allowing the model to explore a diverse set of reasoning paths and solution strategies. The result is a scalable and flexible pipeline for synthesizing varied Code CoTs.

As illustrated in Figure~\ref{fig:case}, even for problem types the model has seen during training—such as calculating the expected value of a biased die—the model is capable of \textit{\textbf{restructuring the logic}}, e.g., by decomposing the problem into multiple rolls and aggregating expected outcomes.
This demonstrates that \texttt{CodeGen} supports two complementary modes of augmentation:

\textbf{\textit{\textcolor{augProblemText}{Problem-level Augmentation}}} arises when natural language problems are synthesized from code by varying the situational context or rephrasing the same underlying logic in different stylistic forms. This introduces diversity in surface formulations (implemented in Section~\ref{met:reverse}).

\textbf{\textit{\textcolor{augPatternText}{Pattern-level Augmentation}}} arises when the CodeGen explores novel reasoning structures—such as problem decompositions or alternative solution strategies—thereby enriching the pool of underlying logic templates.

Together, these modes yield both surface-level diversity and deeper structural variability in the synthesized dataset. Additional representative samples along with training settings and sampling configurations are provided in Appendix~\ref{apx:datagen}.

\noindent{\textbf{Code CoT Filtration.}}
To ensure the quality of generated Code CoTs, we apply the execution-based filtering criteria similar to the unified seed Code CoTs.
The only difference is that at this stage, we do not enforce output matching with known answers, as consistency verification is deferred to the later back-translation stage.
In total, we synthesize approximately 5.3M Code CoT samples. After filtering, we retain a high-quality subset of around 4.6M executable and structurally valid programs.
This large-scale dataset forms the basis for the subsequent stage of problem synthesis, enabling us to bootstrap new question–solution pairs and further expand the reach of code-based reasoning.

\subsection{Instruction Reversal and Language CoT Generation}
\label{met:reverse}
Following the generation of a substantial set of executable code templates, we distill their underlying logic to synthesize natural language problems alongside their corresponding solutions, derived from the combined set $\mathcal{C}_{\text{seed}} \cup \mathcal{C}_\text{samp}$.
As shown in Figure~\ref{fig:case}, this process significantly expands our dataset by producing diverse and high-quality problem-answer pairs.

\noindent{\textbf{Two-Stage QA Generation.}}
For quality control, we adopt a two-stage method to generate problem and language CoT instead of one-step generation for both. 
In our preliminary experiments, we find that jointly generating the instruction and language CoT together based on the Code CoT is easy to lead to low quality or incorrect language solutions, perhaps due to the `lazy' mode~\cite{openmathinstruct} by LLMs since it sees the correct Code CoT as `guidance'. 
Therefore, each code snippet is paired with representative input-output examples (code-instruction pair) and provided as input to the LLM (see Prompt~\ref{pmt:code2math_q}), which generates a natural language problem at the first stage. Secondly, we prompt the generated problem to the LLM (see Prompt~\ref{pmt:code2math_a}) for natural language CoT synthesis, which largely forces the LLM to think and generate correct language CoTs.

\noindent{\textbf{Dual Verification.}}
Two filtration and verification ways are processed to ensure the  correctness of the instruction and natural language CoT. 

(1) \textit{Answer Consistency}: We execute the code and compare its output to the answer inferred from the LLM's CoT reasoning. Any mismatches are discarded to maintain high precision. 

(2) \textit{CoT Consistency}: We remove samples where the language CoT and Code CoT for the same problem are not aligned, based on the consistency judgment in Prompt~\ref{pmt:consist}. This process ensures the correctness of the reasoning steps in the language CoT.

Only tuples $(p', s', c')$ that simultaneously satisfy both conditions are retained. This filtering process can be formally expressed as:
\begin{equation}
\label{eq:final-set}
\mathcal{D}_{\text{final}}
=\{(p',s',c') \mid 
p'=G_{c \to p}(c'),\;
s'=G_{p \to s}(p'),\;
(\mathrm{Ans}(s')=\mathrm{Exec}(c')) \wedge \mathrm{Con}(s',c')\},
\end{equation}
where $G$ denotes a general-purpose LLM used for instruction reversal and answer generation; $\mathrm{Ans}(s')$ extracts the final answer from the solution $s'$ and compares it with the execution result of the code $\mathrm{Exec}(c')$; and $\mathrm{Con}(s', c')$ represents the CoT consistency check between the natural language solution and the code. After this pipeline, we obtain approximately $1.3\text{M}$ validated instruction-answer pairs in $\mathcal{D}_{\text{final}}$, which significantly enhance the diversity and reliability of the training data and serve as a valuable resource for downstream reasoning tasks.

\section{Experiment}
\subsection{Experimental Setup}
\label{exp:setup}

\noindent{\textbf{Baselines.}}
We compare our \method{}-generated dataset against several mainstream synthesized instruction-tuning datasets for math reasoning, including data-centric methods such as MetaMath~\citep{metamath}, MMIQC~\citep{liu2024mmiqc}, NuminaMath~\citep{numinamath}, MathFusion~\citep{mathscale}, RFT~\citep{dartmath}, and DART‑Math~\citep{dartmath}, which all demonstrate strong reasoning enhancement.  
Besides, we also include well-known open-source instruction-tuned or reinforcement learning (RL)-based models as baselines: LLaMA3-7B-Instruct~\citep{llama3}, Qwen2.5-Math-Instruct~\citep{qwen25_math}, and DeepSeekMath-7B-RL~\citep{deepseekmath}.

\noindent{\textbf{Training Configuration.}}
To evaluate the generalizable effectiveness of our \texttt{Caco} produced dataset,
our experiments are conducted on two math-specialized LLMs—DeepSeekMath‑7B~\cite{deepseekmath} and Qwen2.5‑Math‑7B~\cite{qwen25_math}, as well as one general-purpose model, LLaMA3‑8B~\cite{llama3}.  
Unless otherwise specified, all models are fine-tuned for 3 epoch using a learning rate of $5\times10^{-6}$, a batch size of 128, and a cosine decay schedule with a warm-up ratio of 0.03.  
Additional implementation details are provided in Appendix~\ref{apx:train_detail}.

\noindent{\textbf{Evaluation Setup.}}
Following the evaluation protocol of DartMath~\citep{dartmath}, we evaluate on multiple popular benchmarks to show the advantages, including MATH~\citep{hendrycks2021math}, GSM8K~\citep{cobbe2021gsm8k}, CollegeMath~\citep{mathscale}, DeepMind‑Mathematics~\citep{saxton2018dmmath}, OlympiadBench‑Math~\citep{he2024olympiadbench}, and TheoremQA~\citep{chen2023theoremqa}.  
Solutions are generated using greedy decoding with a maximum sequence length of 2048 tokens, and we report Pass@1 accuracy in the zero-shot setting without tool integration.  
Further evaluation details and benchmark statistics can be found in Appendix~\ref{apx:evaluation_detail}.

\subsection{Main Results}
\label{exp:main}
\begin{table}[!htbp]
    \vspace{-0.5cm}
    \centering
      \resizebox{\linewidth}{!}{
        \begin{tabular}{lccccccccc}
          \toprule
          \multicolumn{1}{c}{\textbf{Model}} & \textbf{\# Samples} & \textbf{MATH} & \textbf{GSM8K} & \textbf{College} & \textbf{DM} & \textbf{Olympiad} & \textbf{Theorem} & \textbf{AVG} \\
          \midrule
          \multicolumn{10}{c}{\textbf{\textit{DeepSeekMath-7B (Math-Specialized Base Model)}}} \\
          \midrule
DeepSeekMath-7B-RL         & -    & 51.1 & \textbf{88.8} & 34.5 & 58.2 & 18.8 & 30.9 & 47.1 \\
DeepSeekMath-7B-MetaMath   & 400K & 40.2 & 80.5 & 35.7 & 48.1 & 11.4 & 21.8 & 39.6  \\
DeepSeekMath-7B-MMIQC$^\dagger$      & 2.3M & 45.3 & 79.0 & 35.3 & 52.9 & 13.0 & 23.4 & 41.5  \\
DeepSeekMath-7B-NuminaMath & 860K & 47.7 & 78.5 & 38.0 & 56.2 & 18.2 & 22.1 & 43.5  \\
DeepSeekMath-7B-RFT$^\dagger$        & 590K & 53.0 & 88.2 & 41.9 & 60.2 & 19.1 & 27.2 & 48.3  \\
DeepSeekMath-7B-DartMath$^\dagger$   & 590K & 53.6 & 86.8 & 40.7 & 61.6 & 21.7 & 32.2 & 49.4  \\
DeepSeekMath-7B-MathFusion$^\dagger$ & 60K  & 53.4 & 77.9 & 39.8 & 65.8 & 23.3 & 24.6 & 47.5  \\
\rowcolor[rgb]{ .867, .922, .969} {\texttt{Caco}-Seed109K-DeepSeekMath-7B} & 109K & 58.7 & 82.4 & 42.9 & 71.3 & 22.4 & 28.9 & 51.1 \\
\rowcolor[rgb]{ .867, .922, .969} {\texttt{Caco}-596K-DeepSeekMath-7B}  & 596K & 63.5 & 85.2 & 44.4 & 78.0 & 25.8 & 30.2 & 54.5  \\
\rowcolor[rgb]{ .867, .922, .969} {\texttt{Caco}-1.3M-DeepSeekMath-7B}  & 1.3M & \textbf{68.2} & 85.1 & \textbf{46.0} & \textbf{80.2} & \textbf{29.5} & \textbf{33.8} & \textbf{57.1} \\
          \midrule
          \multicolumn{10}{c}{\textbf{\textit{Qwen2.5-Math-7B (Math-Specialized Base Model)}}} \\
          \midrule
Qwen2.5-Math-7B-Instruct   & -    & 82.1 & \textbf{94.1} & 50.4 & 72.9 & 41.3 & 40.8 & 63.6 \\
Qwen2.5-Math-7B-MetaMath   & 400K & 51.7 & 84.7 & 40.0 & 62.6 & 18.2 & 26.5 & 47.3 \\
Qwen2.5-Math-7B-NuminaMath & 860K & 70.6 & 90.8 & 46.1 & 75.1 & 35.9 & 37.4 & 59.3 \\
Qwen2.5-Math-7B-DartMath   & 590K & 61.4 & 89.7 & 42.5 & 72.0 & 25.8 & 35.5 & 54.5 \\
Qwen2.5-Math-7B-MathFusion & 60K  & 75.2 & 83.5 & 43.0 & 76.0 & 39.5 & 41.5 & 59.8 \\
\rowcolor[rgb]{ .867, .922, .969} {\texttt{Caco}-Seed109K-Qwen2.5-Math-7B} & 109K & 80.6 & 92.3 & 47.1 & 83.0 & 41.6 & 45.9 & 65.1 \\
\rowcolor[rgb]{ .867, .922, .969} {\texttt{Caco}-596K-Qwen2.5-Math-7B}  & 596K & 81.1 & 92.4 & 50.3 & 86.7 & 43.3 & 45.5 & 66.6 \\
\rowcolor[rgb]{ .867, .922, .969} {\texttt{Caco}-1.3M-Qwen2.5-Math-7B}  & 1.3M & \textbf{82.4} & 92.6 & \textbf{51.4} & \textbf{87.1} & \textbf{46.5} & \textbf{46.0} & \textbf{67.7} \\
          \midrule
          \multicolumn{10}{c}{\textbf{\textit{LLaMA3-8B (General Base Model)}}} \\
          \midrule
LLaMA3-8B-Instruct   & -    & 44.3 & 53.4 & 29.8 & 42.0 & 11.3 & 17.7 & 33.1 \\
LLaMA3-8B-MetaMath$^\dagger$   & 400K & 32.5 & 77.3 & 20.6 & 35.0 & 5.5  & 13.8 & 30.8 \\
LLaMA3-8B-MMIQC$^\dagger$      & 2.3M & 39.5 & 77.6 & 29.5 & 41.0 & 9.6  & 16.2 & 35.6 \\
LLaMA3-8B-NuminaMath & 860K & 43.6 & 79.7 & 24.7 & 43.1 & 16.4 & 19.9 & 37.9 \\
LLaMA3-8B-RFT$^\dagger$        & 590K & 39.7 & 81.7 & 23.9 & 41.7 & 9.3  & 14.9 & 35.2 \\
LLaMA3-8B-DartMath$^\dagger$   & 590K & 46.6 & 81.1 & 28.8 & 48.0 & 14.5 & 19.4 & 39.7 \\
LLaMA3-8B-MathFusion$^\dagger$ & 60K  & 46.5 & 79.2 & 27.9 & 43.4 & 17.2 & 20.0 & 39.0 \\
\rowcolor[rgb]{ .867, .922, .969} {\texttt{Caco}-Seed109K-Llama3-8B}       & 109K & 55.3 & 86.0 & 42.2 & 52.0 & 19.1 & 25.6 & 46.7 \\
\rowcolor[rgb]{ .867, .922, .969} {\texttt{Caco}-596K-LLaMA3-8B}  & 596K & 64.3 & 88.6 & 44.8 & 66.7 & 24.7 & 27.6 & 52.8 \\
\rowcolor[rgb]{ .867, .922, .969} {\texttt{Caco}-1.3M-LLaMA3-8B}  & 1.3M & \textbf{70.6} & \textbf{89.1} & \textbf{46.2} & \textbf{72.5} & \textbf{34.1} & \textbf{31.0} & \textbf{57.3} \\
          \bottomrule
        \end{tabular}
    }
      \caption{
        Performance comparison on mathematical benchmarks including MATH, GSM8K, CollegeMATH (College), DeepMind-Mathematics (DM), OlympiadBench-Math (Olympiad), and TheoremQA (Theorem). 
        The best results are highlighted in \textbf{bold}.
        Baseline results labeled with $^\dagger$ are derived from 
 MathFusion~\citep{pei2025mathfusion}.}
  \label{tab:main_result}
  \vspace{-0.5cm}
\end{table}

Table~\ref{tab:main_result} presents a comprehensive comparison of our \texttt{Caco} against a series of strong baselines across the three different base models (DeepSeekMath-7B, Qwen2.5-Math-7B, and LLaMA3-8B). We report results for two synthesized data sizes: \texttt{Caco}-596K and \texttt{Caco}-1.3M samples. From the results, we can summarize the following findings:

\textbf{Consistent improvements across base models.} \texttt{Caco} consistently outperforms existing methods across all three base models. 
For instance, on LLaMA3-8B, \texttt{Caco}-1.3M achieves an average score of 57.3, surpassing the previous best of 39.7 from DartMath~\cite{dartmath} by a relative improvement of 44.3\%.
    
\textbf{Improvement over scaled synthetic data.} Performance improves obviously when increasing the \texttt{Caco}-generated data from 596K to 1.3M. On Qwen2.5-Math-7B, \texttt{Caco}-1.3M achieves 67.7, outperforming \texttt{Caco}-596K by 1.1 and demonstrating the scalability and effectiveness of our approach under larger supervision.

\textbf{Strong performance on challenging subsets.} Notably, \texttt{Caco} shows superior performance on harder benchmarks such as OlympiadBench and TheoremQA, where other baselines struggle. For instance, on LLaMA-8B, \texttt{Caco}-596K improves OlympiadBench from 17.2 to 34.1 and TheoremQA from 20.0 to 31.0 compared to MathFusion, which shows the great potential of our approach.

\textbf{Competitive with strong instruction-tuned and RL-based models.} Remarkably, \texttt{Caco} matches or exceeds the performance of strong instruction-tuned or RL-finetuned models. For example, on Qwen2.5-Math-7B, \texttt{Caco}-1.3M achieves 67.7, which is comparable to Qwen2.5-Math-7B-Instruct (63.6). On DeepSeekMath and LLaMA series, \texttt{Caco}-1.3M trained models significantly surpass DeepSeekMath-7B-RL (47.1) and LLaMA-8B-Instruct (33.1). This greatly demonstrates the superiority of our method.

\textbf{Effectiveness of \texttt{Caco} Data.} Compared to the seed data we used to train \texttt{CodeGen} (\texttt{Caco}-Seed-109K), \texttt{Caco}-596K and \texttt{Caco}-1.3M consistently deliver substantial improvements. For instance, on LLaMA3-8B, \texttt{Caco}-1.3M achieves 57.3, a significant increase from \texttt{Caco}-Seed-109K’s score of 46.7. This validates our data scaling strategy, showing that our method yields performance gains by ensuring the training data comprehensively represents diverse and challenging problems.

\section{Analysis}
To further understand the strengths of our proposed approach, we analyze three key aspects that contribute to \texttt{Caco}'s effectiveness: the \textit{diversity}, the \textit{scalability}, and the \textit{verification} mechanism in the \texttt{Caco} data construction pipeline. Together, these components form the foundation of \texttt{Caco}'s training methodology and help explain its strong performance across models and benchmarks. In the following sections, we provide a detailed analysis of each component and its contribution. More experiments and discussion of cost are in Appendix~\ref{apx:more_exp} and ~\ref{apx:cost}.

\subsection{Analysis on Data Diversity}
We conduct a comprehensive investigation into the diversity of the dataset to assess the range and variability of the \texttt{Caco}-generated problems. This analysis is crucial for understanding how well the model can generate problems across various domains and ensure broad coverage of mathematical topics. By examining both the distribution of problems and the variety of problem types, we aim to demonstrate that the dataset not only spans a wide range of topics but also captures diverse problem-solving scenarios that are representative of real-world mathematical challenges.

\begin{figure}[htbp]
    \vspace{-0.3cm}
    \begin{minipage}{\linewidth}
    \begin{minipage}{\textwidth}
            \centering
    \begin{subfigure}[b]{0.3\linewidth}
        \centering
        \includegraphics[height=4cm]{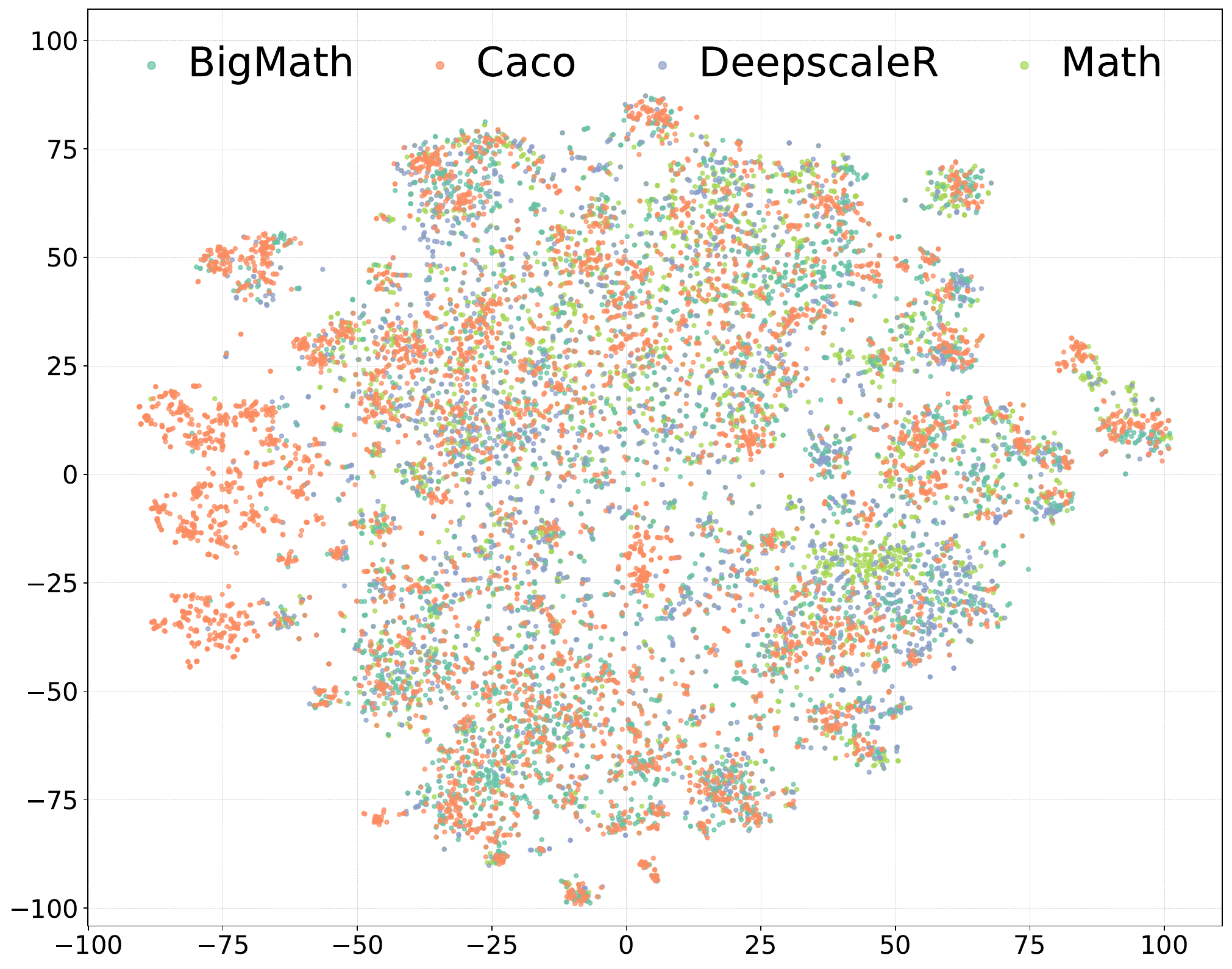}
        \phantomcaption
        \label{fig:dist}
        \vspace{-0.3cm}
    \end{subfigure}
    \hspace{0.1\linewidth}
    \begin{subfigure}[b]{0.4\linewidth}
        \centering
        \includegraphics[height=4cm]{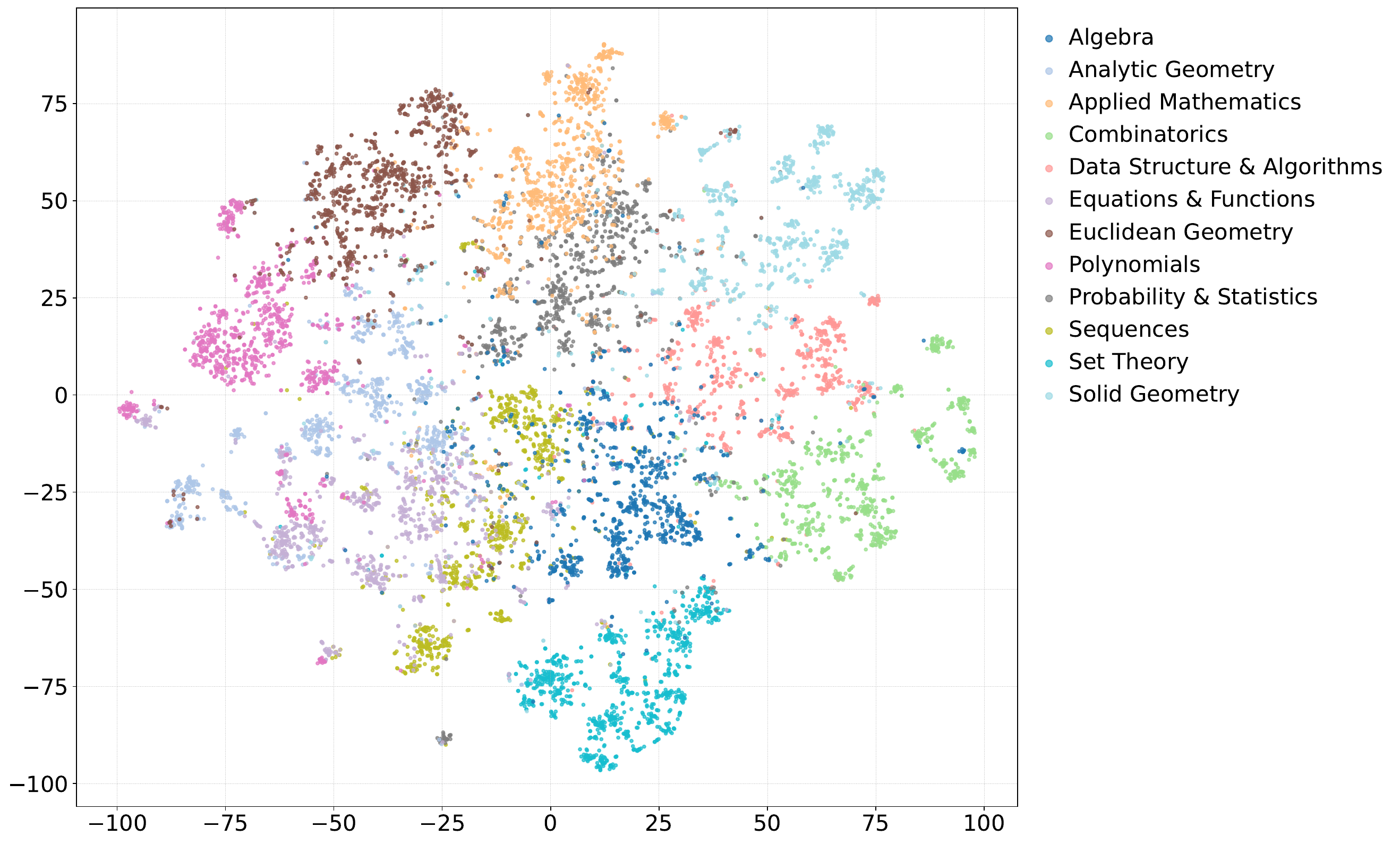}
        \phantomcaption
        \label{fig:cluster}
        \vspace{-0.3cm}
    \end{subfigure}
    \end{minipage}
    \end{minipage}
    \caption{\textbf{Left}: Problem distribution of our \texttt{Caco} dataset and the original data sources.
    \textbf{Right}: KMeans clustering result of the problem types.}
    \label{fig:all}
    \vspace{-0.5cm}
\end{figure}

\noindent{\textbf{Problem Diversity.}}
We analyze the distribution of problems in the synthesized \texttt{Caco} dataset to assess its coverage and diversity. Specifically, we randomly sample 5K problems from \texttt{Caco} and compare them with samples from the original seed datasets (MATH, DeepScaleR and BigMath). We encode all problems using the all-MiniLM-L6-v2 sentence embedding model\footnote{\url{https://huggingface.co/sentence-transformers/all-MiniLM-L6-v2}}, and visualize their distributions via t-SNE~\cite{tsne}, as shown in Figure~\ref{fig:dist}.
The resulting plot demonstrates that \texttt{Caco}'s synthesized data broadly and evenly spans the embedding space, effectively covering the original seed distributions. Notably, we observe a distinct region on the left side of the plot where \texttt{Caco} samples diverge from the seed data clusters, suggesting that our generation pipeline introduces novel and diverse problem types beyond the original datasets. This supports the claim that \texttt{Caco} enhances distributional generalization through its diverse synthetic augmentation.

\noindent{\textbf{Topic Diversity.}}
To further assess the topical diversity of the \texttt{Caco} dataset, we apply clustering analysis to the problem embeddings. Using the same embedding method as before, we encode all problems and then apply the KMeans algorithm~\cite{kmeans} to partition them into 12 distinct clusters. The clustering results are visualized in Figure~\ref{fig:cluster}.
The clusters reveal a wide range of mathematical and algorithmic topics, including algebra, geometry, applied mathematics, data structures, algorithms, and more. This confirms that \texttt{Caco} spans a broad spectrum of problem types, rather than concentrating on narrow domains. Representative samples from each identified topic cluster are provided in Appendix~\ref{apx:cases} for qualitative reference.

\subsection{Analysis on Data Scalability}

We evaluate the scalability of \texttt{Caco} by analyzing its impact on model performance under varying amounts of training data. Figure~\ref{fig:scale} presents the results on the MATH benchmark and the overall average across all benchmarks for DeepSeekMath-7B and LLaMA3-8B.
For both models, we observe a clear upward trend as the training data size increases from 109K to 1.3M. This demonstrates the strong scalability of our approach. Notably, the performance gains are more pronounced for the general-purpose LLaMA3-8B, especially in the early stages (e.g., from Seed109K to 596K), highlighting \texttt{Caco}'s ability to significantly improve less specialized models. On Qwen2.5-Math model, the performance also improves with increasing data size, but the improvement is less pronounced due to the already strong capabilities of the base model.

\subsection{Ablation on Verification}

\begin{figure}[htbp]
    \vspace{-0.3cm}
    \begin{minipage}{\linewidth}
    \begin{minipage}{\textwidth}
            \centering
    \begin{subfigure}[b]{0.3\linewidth}
        \centering
        \includegraphics[height=4cm]{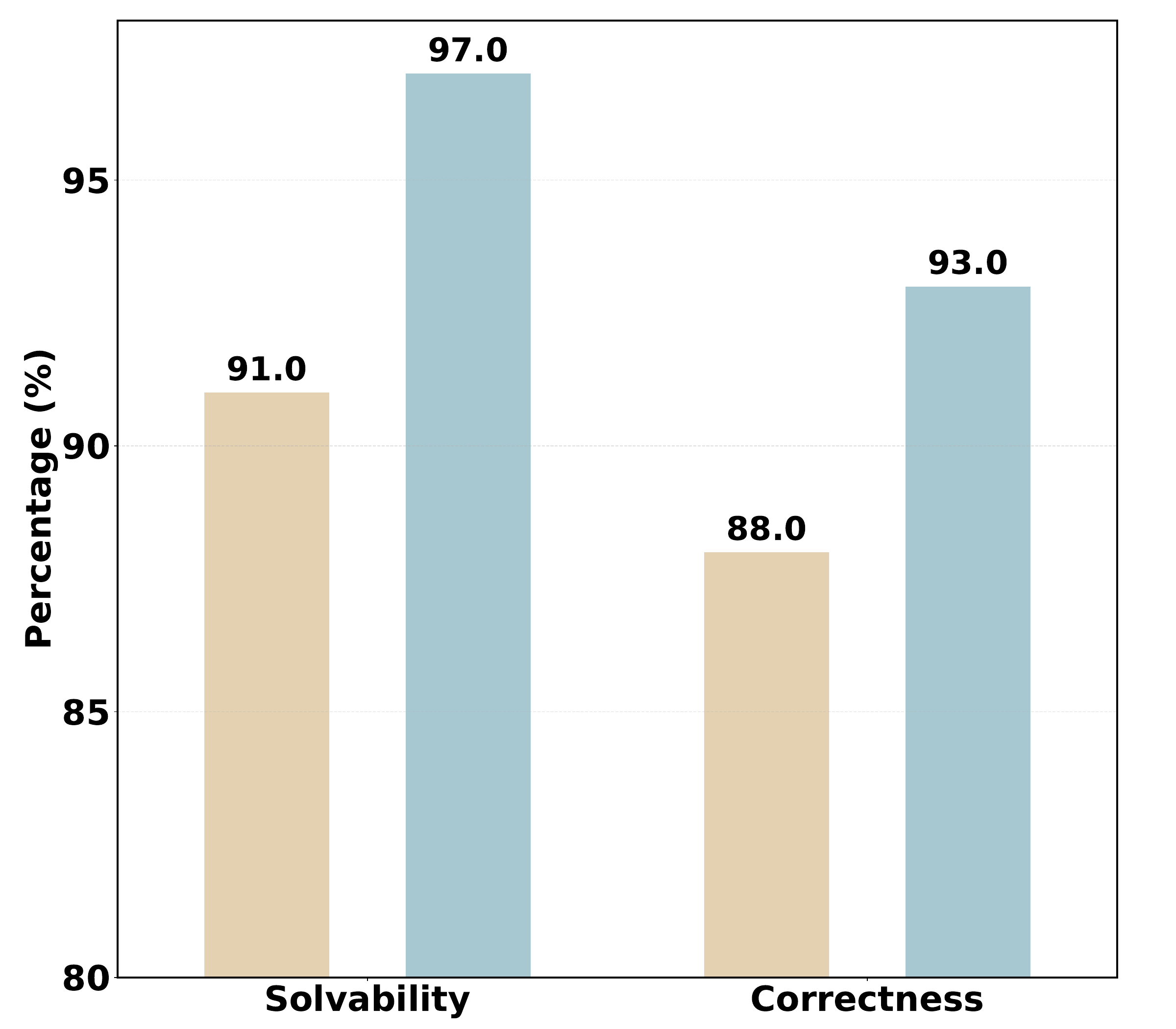}
        \phantomcaption
        \label{fig:quality}
        \vspace{-0.3cm}
    \end{subfigure}
    \hfill
    \begin{subfigure}[b]{0.3\linewidth}
        \centering
        \includegraphics[height=4cm]{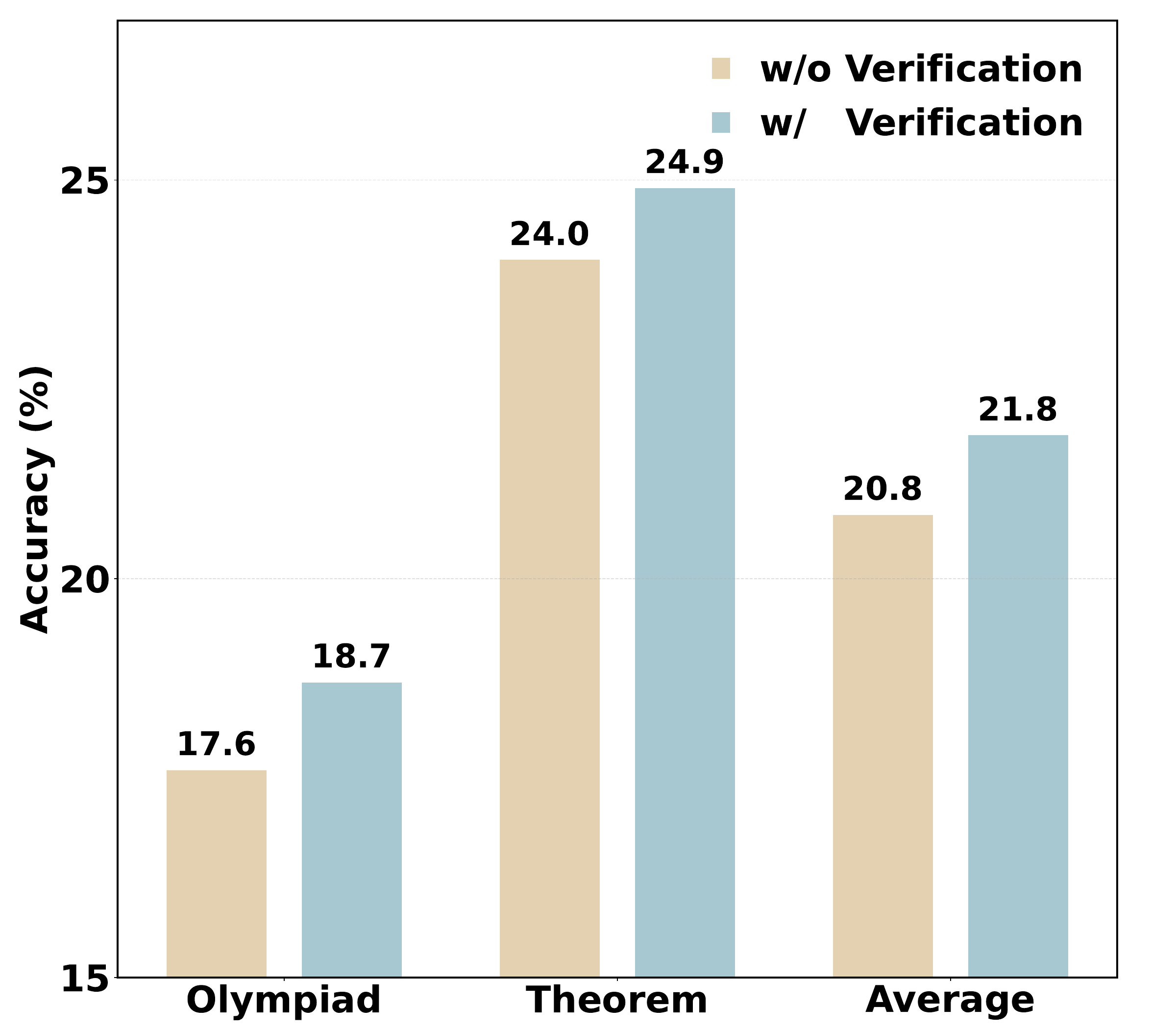}
        \phantomcaption
        \label{fig:model}
        \vspace{-0.3cm}
    \end{subfigure}
    \hfill
    \begin{subfigure}[b]{0.3\linewidth}
        \centering
        \includegraphics[height=4cm]{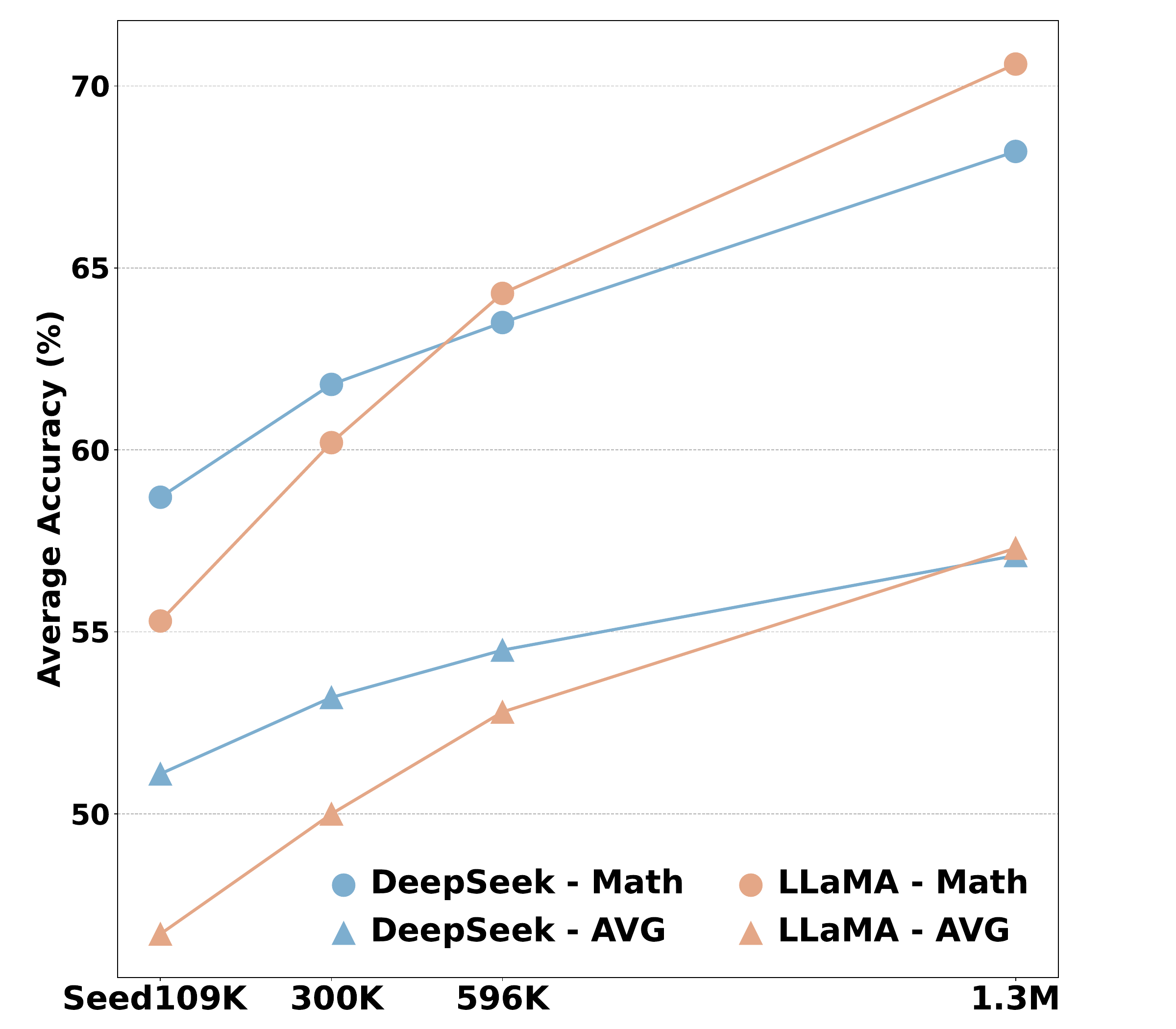}
        \phantomcaption
        \label{fig:scale}
        \vspace{-0.3cm}
    \end{subfigure}
    \end{minipage}
    \end{minipage}
    \caption{\textbf{Left}: Comparison of solvability and correctness between generated samples with and without verification.
    \textbf{Middle}: Accuracy comparison between models trained on verified and non-verified data. 
    \textbf{Right}: Performance improvements of the \texttt{Caco} model as data size increases.}
    \label{fig:all}
    \vspace{-0.3cm}
\end{figure}
Verification is a crucial process in our \texttt{Caco} data generation. To further investigate the impact of verification on data quality and reliability, we compare the data with and without applying the verification filtering. We randomly sample 100K data points for each version and use Qwen3-32B to evaluate both the solvability (i.e., whether the problem can be solved) and the correctness (i.e., whether the final answer is accurate) of the generated samples. We further evaluate downstream performance by fine-tuning the LLaMA model on each dataset.

\noindent{\textbf{Impact on Data Quality.}}
As shown in Figure~\ref{fig:quality}, the verification mechanism substantially improves the quality of the training data. With verification, the ratio of solvable problems increases from 91K to 97K, and the number of correct answers rises from 88K to 93K. These improvements suggest that the verification process—based on answer validation and consistency checks over reasoning chains—effectively filters out low-quality or incorrect samples, resulting in more reliable supervision.

\noindent{\textbf{Impact on Model Performance.}}
In addition to improving data quality, verification also yields tangible benefits in downstream performance (Figure~\ref{fig:model}). The model trained with verified data achieves an average accuracy of 21.8, compared to 20.8 without verification, reflecting a consistent improvement across benchmarks. The performance gain is especially notable on more challenging tasks: for instance, on Olympiad, the verified model scores 18.7, outperforming its non-verified counterpart by 1.1 points. This demonstrates that the enhanced data reliability introduced by verification translates into better generalization and reasoning robustness in trained models.

\subsection{Generality Beyond Mathematics} 
\label{ana:general}
We first evaluate the generalization of \texttt{Caco} models. Using OpenCompass~\cite{2023opencompass}, we assess \texttt{Caco}-1.3M models across a broad set of reasoning tasks, including mathematics (AIME24), code generation (HumanEval+), scientific QA (ARC-c), logic puzzles (BBH, KorBench), and general knowledge/science (AGIEval). 
\texttt{Caco} models demonstrate substantial improvements beyond math, with notable gains in logic puzzles, general reasoning, science reasoning, and code tasks. 
These results indicate that the models trained on \texttt{Caco} data generalize effectively across diverse benchmarks.

\begin{table}[!htbp]
    \centering
    \resizebox{\linewidth}{!}{
        \begin{tabular}{lccccccc}
            \toprule
            \multicolumn{1}{c}{\textbf{Model}} & \textbf{AGIEval} & \textbf{AIME24} & \textbf{HumanEval+} & \textbf{ARC-c} & \textbf{BBH} & \textbf{KorBench} & \textbf{Average} \\
            \midrule
            Qwen2.5-Math-7B-base & 42.5 & 20.0 & 12.8 & 72.2 & 19.9 & 39.7 & 34.5 \\
            \texttt{Caco}-1.3M-Qwen2.5-Math-7B & \textbf{53.3} & \textbf{23.3} & \textbf{53.1} & \textbf{81.4} & \textbf{65.1} & \textbf{47.1} & \textbf{53.9} \\
            \midrule
            LLaMA3-8B-base & 28.5 & 0.0 & 32.3 & 79.0 & 19.8 & 23.8 & 30.6 \\
            \texttt{Caco}-1.3M-LLaMA3-8B & \textbf{46.5} & \textbf{10.8} & \textbf{34.2} & \textbf{83.1} & \textbf{33.8} & \textbf{44.1} & \textbf{42.1} \\
            \bottomrule
        \end{tabular}
    }
    \caption{Performance comparison of base models and \texttt{Caco}-augmented models across diverse out-of-domain benchmarks.}
    \label{tab:caco_generalization}
\end{table}

\begin{table}[!htbp]
    \vspace{-0.3cm}
    \centering
    \resizebox{0.8\textwidth}{!}{
        \begin{tabular}{lcccc}
            \toprule
            \multicolumn{1}{c}{\textbf{Model}} & \textbf{AGIEval} & \textbf{ARC-c} & \textbf{MMLU-STEM} & \textbf{Average} \\
            \midrule
            LLaMA-MegaScience-Seed5.2K & 42.8 & 78.6 & 55.4 & 59.0 \\
            LLaMA-MegaScience-Caco37K  & \textbf{45.0} & \textbf{84.8} & \textbf{60.5} & \textbf{63.4} \\
            \bottomrule
        \end{tabular}
    }
    \caption{Evaluation of LLaMA models trained on MegaScience seed data (5.2K) vs. \texttt{Caco}-augmented expansion (37K).}
    \label{tab:megasci_generalization}
\end{table}

We next discuss the generality of the \texttt{Caco} methodology itself beyond mathematics. 
Although our primary experiments focused on mathematical reasoning, \texttt{Caco} is fundamentally a general-purpose framework for structured, code-based reasoning, and is applicable to domains exhibiting \emph{logical, symbolic, or programmatic} structure, such as logic puzzles, scientific reasoning, and procedural tasks. 
In logic puzzles, for instance, many problems share a reusable underlying reasoning template (e.g., arithmetic expression puzzles, countdown problems), which can be parameterized in code to generate diverse instances. This aligns with \texttt{Caco}'s central principle: \emph{code abstracts problem logic more compactly than natural language}, enabling systematic sampling and verification.

To test cross-domain applicability, we applied \texttt{Caco} to 5.2K science reasoning seeds from MegaScience. The pipeline generated 37K valid QA samples, and fine-tuning LLaMA on these yielded an average score improvement from 59.0 to 63.4 across AGIEval, ARC-c, and MMLU-STEM (Table~\ref{tab:megasci_generalization}). These results confirm that \texttt{Caco}'s code-driven design enables effective extension to new domains where logic can be programmatically represented.

\section{Conclusion}
In this work, we present \texttt{Caco}, a code-assisted framework for generating high-quality, verifiable, and diverse chain-of-thought reasoning data. By leveraging code execution and automated filtering, \texttt{Caco} enables scalable synthesis of logically grounded instruction data without human supervision. Models trained with \texttt{Caco} outperform strong baselines on both mathematical reasoning benchmarks and out-of-domain benchmarks. Our findings highlight the effectiveness of code-driven verification and instruction diversity in improving reasoning generalization.
\clearpage
\bibliographystyle{mybst}
\bibliography{ref}

\clearpage
\appendix
\section{Data Generation}
\label{apx:datagen}
\subsection{Prompts}
We show the prompts used for Code CoT unifying in Prompt \ref{pmt:math2code} and Prompt \ref{pmt:unify_code}, and CodeGen training and sampling in Prompt \ref{pmt:codegen_train} and \ref{pmt:codegen_sample}. We also provide the back-translation prompt for question generation in Prompt \ref{pmt:code2math_q} and answer generation prompt in Prompt \ref{pmt:code2math_a}.
prompts for problem solvability, answer correctness, the consistency between the answer and the code's chain-of-thought (CoT) evaluations are displayed in Prompt~\ref{pmt:solvable},~\ref{pmt:correct}, and~\ref{pmt:consist}.

\subsection{Model Usage}
We detail the models employed at each stage of our pipeline:
\begin{itemize}
\item \textbf{Unifying Code CoTs:} We used Qwen2.5-72B-Instruct to generate unified Code CoTs.
\item \textbf{CodeGen:} The unconditional CodeGen was fine-tuned from Qwen2.5-Coder-7B.
\item \textbf{Problem Reversal \& Solution Generation:} Both question back-translation and answer synthesis were performed using Qwen3-8B.
\item \textbf{Evaluation:} All assessments (problem solvability, answer correctness, and CoT consistency) were conducted with Qwen3-32B.
\end{itemize}

\subsection{Implementation Details}
\label{apx:imp_detail}
\noindent{\textbf{Focusing on Challenging Code CoT.}}
To increase the difficulty of the dataset, we applied additional filtering to the largest subset, bigmath, of the Code CoT dataset. Based on the solve rate annotations provided with the dataset, we retained only those Code CoTs with a solve rate of less than 0.3.

\noindent{\textbf{Hyperparameters.}}
During the Unifying Code CoT stage, we deployed Qwen2.5-72B-Instruct on 4 A100 GPUs to generate code from the raw datasets. For each sample, we performed a single pass of sampling with a temperature of 0.6.

For training the CodeGen model, we used the LlamaFactory framework and adopted the same training configuration as in the main experiments. During inference, we sampled with a temperature of 0.9 and a maximum sequence length of 1024 tokens.

For problem and solution generation, we followed the Qwen3-8B best practice~\citep{qwen3}. Specifically, we used: \texttt{Temperature = 0.7, TopP = 0.8, TopK = 20, MinP = 0}, and \texttt{enable\_thinking = False}.  

We use Qwen3-32B for evaluating problem solvability, answer correctness, and the CoT consistency between the natural language solution and code CoT.

\subsection{Filtering Mechanism for Code CoTs}
\label{method:filter}
As discussed in method section, many stages of our pipeline require rigorous filtering to ensure the quality, correctness, and executability of the generated Code CoTs. Here, we formally describe the filtering criteria used throughout our work.

\begin{itemize}
    \item \textbf{Executability.} The code must be syntactically valid and executable without raising runtime errors. This ensures basic correctness and structural integrity.
    \item \textbf{Execution Efficiency.} To prevent degenerate or non-terminating programs, we discard any samples that exceed a 10-second execution time limit under a controlled runtime environment.
    \item \textbf{Minimum Code Length.} To avoid trivial or underdeveloped solutions, we require that each code snippet contain at least six non-comment lines of code. This encourages a minimal degree of reasoning complexity and explanatory depth.
    \item \textbf{AST-Based Semantic Validation.} Using abstract syntax tree (AST) analysis~\cite{ast}, we ensure that all variables declared in the input dictionary are functionally utilized in the program’s logic. This discourages redundant or templated outputs and promotes semantically meaningful solutions.
    \item \textbf{Output Consistency.} When ground-truth answers are available, we verify that the program output exactly matches the expected solution. This check is applied in cases where reference answers are known and consistency can be reliably evaluated.
\end{itemize}

\section{Train and Evaluation}
\label{apx:train_detail}
\label{apx:evaluation_detail}
\subsection{Training Setup}
Model training was conducted using the LLaMA Factory~\footnote{\url{https://github.com/hiyouga/LLaMA-Factory}} framework on 8 NVIDIA A100 GPUs.  
All models were trained for 3 epoch with a batch size of 128. We used the AdamW optimizer~\citep{adamw} with a learning rate of $5 \times 10^{-6}$, cosine learning rate decay, and a warm-up ratio of 0.03.  
The maximum sequence length (cutoff) was set to 4096, and the weight decay was 0.1.  
The prompt used for training is shown in Prompt~\ref{pmt:train_prompt}.

\subsection{Evaluation Setup}
All models were evaluated using a unified framework~\footnote{\url{https://github.com/ZubinGou/math-evaluation-harness/tree/main}} under the zero-shot setting. We used greedy decoding with a maximum generation length of 2048 tokens.
The prompt used for evaluation is shown in Prompt~\ref{pmt:eval_prompt}.

\subsection{Evaluation Benchmarks}
The following datasets are used for evaluation:

\begin{itemize}
  \item \textbf{MATH}~\citep{hendrycks2021math}: A benchmark of 12,500 high school math competition problems, with 7,500 for training and 5,000 for testing. Problems are categorized into 7 topics (Prealgebra, Intermediate Algebra, Algebra, Precalculus, Geometry, Counting \& Probability, and Number Theory) and 5 difficulty levels.

  \item \textbf{GSM8K}~\citep{cobbe2021gsm8k}: This dataset contains 8,792 high-quality grade school math word problems, with 7,473 for training and 1,319 for testing. Each problem typically requires 2 to 8 reasoning steps to solve.

  \item \textbf{CollegeMath}~\citep{mathscale}: A test set containing 2,818 college-level math problems collected from 9 college textbooks, covering 7 core subjects: Algebra, Precalculus, Calculus, Vector Calculus, Probability, Linear Algebra, and Differential Equations.

  \item \textbf{DeepMind-Mathematics}~\citep{saxton2018dmmath}: This test set consists of 1,000 problems covering a wide range of mathematical reasoning tasks including algebra, arithmetic, calculus, and probability. It is designed to assess the mathematical reasoning abilities of models.

  \item \textbf{OlympiadBench-Math}~\citep{he2024olympiadbench}: A benchmark of 675 Olympiad-level math problems. We evaluate only on the English text-only subset of OlympiadBench.

  \item \textbf{TheoremQA}~\citep{chen2023theoremqa}: A theorem-driven question-answering benchmark containing 800 problems grounded in 350 domain-specific theorems. It evaluates a model’s ability to apply mathematical and scientific theorems across disciplines such as mathematics, physics, electrical engineering, computer science, and finance.
\end{itemize}

\section{Additional Experiments}
\label{apx:more_exp}
\subsection{Distinguishing \texttt{\texttt{Caco}} from Teacher Knowledge Transfer and STaR-style Self-Improvement} 
A natural concern is whether \texttt{\texttt{Caco}}’s performance gains stem primarily from knowledge transfer from the large teacher model (Qwen-2.5-72B-Instruct) used to generate the seed dataset, rather than from the \texttt{\texttt{Caco}} procedure itself. 
To isolate this factor, we conducted a control experiment in which the same teacher model was used to directly produce natural language Chain-of-Thought (CoT) answers for the same seed questions, resulting in a 300K QA dataset (\textsc{Qwen72B-Seed-distilled}). 
We compared models fine-tuned on this dataset to those trained on a 300K subset of \texttt{\texttt{Caco}} (\texttt{Caco}-300K) and the full \textsc{\texttt{Caco}-1.3M}. 
Even at equal data size, \texttt{Caco} outperformed the distilled baseline (e.g., 66.2 vs. 65.5 AVG for Qwen-7B), and scaling \texttt{Caco} to 1.3M samples yielded further improvements (up to 67.7 AVG). 
This suggests that prompt and reasoning diversity, enabled by \texttt{Caco}’s code-based augmentation, provides benefits beyond direct teacher distillation, and that \texttt{Caco} scales more effectively.

Another question is whether the gains could also be achieved by simpler self-improvement methods such as STaR~\cite{zelikman2022star}. 
Conceptually, \texttt{Caco} differs from these in that it does not focus on iteratively refining answers to a fixed set of questions; instead, it trains a dedicated code generator to produce executable CoTs from scratch, enabling scalable, verifiable creation of new problems. 
Nevertheless, to provide a direct comparison, we implemented a single iteration of STaR-style self-improvement on the seed dataset, generating multiple CoTs per seed question, filtering for correctness, and sampling $300$K verified solutions (\textsc{Seed-self-improve}). 
Across both DeepSeek-Math-7B and Qwen-7B backbones, \textsc{\texttt{Caco}-300K} consistently outperformed \textsc{Seed-self-improve} by substantial margins (e.g., $53.2$ vs. $48.9$ AVG for DS-7B, and $66.2$ vs. $52.6$ for Qwen-7B). 
These results reinforce that \texttt{Caco}’s improvements derive from its code-driven, diversity-oriented generation process, rather than simply inheriting knowledge from a stronger teacher or applying standard self-improvement on seed data.

\begin{table}[!htbp]
    \centering
    \resizebox{\linewidth}{!}{
        \begin{tabular}{lcccccccc}
            \toprule
            \multicolumn{1}{c}{\textbf{Model}} & \textbf{\#Samples} & \textbf{MATH} & \textbf{GSM8K} & \textbf{College} & \textbf{DM} & \textbf{Olympiad} & \textbf{TheoremQA} & \textbf{AVG} \\
            \midrule
            Qwen-7B-Seed-self-improve      & 300K & 70.7 & 83.0 & 47.1 & 47.6 & 39.0 & 28.2 & 52.6 \\
            Qwen-7B-Qwen72B-Seed-distilled & 300K & 79.0 & 91.2 & 52.1 & 84.4 & 41.3 & 45.0 & 65.5 \\
            Qwen-7B-Caco-300K              & 300K & 81.6 & 92.4 & 51.2 & 84.8 & 42.5 & 44.9 & 66.2 \\
            \textbf{Qwen-7B-Caco-1.3M}     & \textbf{1.3M} & \textbf{82.4} & \textbf{92.6} & \textbf{51.4} & \textbf{87.1} & \textbf{46.5} & \textbf{46.0} & \textbf{67.7} \\
            \midrule
            DS-7B-Seed-self-improve        & 300K & 53.1 & 86.7 & 41.6 & 62.5 & 19.3 & 30.2 & 48.9 \\
            DS-7B-Qwen72B-Seed-distilled   & 300K & 57.4 & 83.0 & 42.4 & 69.0 & 23.4 & 31.4 & 51.1 \\
            DS-7B-Caco-300K                & 300K & 61.8 & 83.2 & 43.3 & 76.0 & 23.9 & 31.1 & 53.2 \\
            \textbf{DS-7B-Caco-1.3M}       & \textbf{1.3M} & \textbf{68.2} & \textbf{85.1} & \textbf{46.0} & \textbf{80.2} & \textbf{29.5} & \textbf{33.8} & \textbf{57.1} \\
            \bottomrule
        \end{tabular}
    }
    \caption{Control experiment comparing teacher-distilled natural language CoTs vs. Caco-generated data at matched size (300K) and at scale (1.3M). Bold indicates the best within each student block.}
    \label{tab:control_distill_vs_caco}
\end{table}

\section{Computational Cost and Efficiency} 
\label{apx:cost}
\begin{table}[!htbp]
    \centering
    \resizebox{0.7\linewidth}{!}{
        \begin{tabular}{lcc}
            \toprule
            \textbf{Stage} & \textbf{\#Samples} & \textbf{Time (Hours)} \\
            \midrule
            Unifying Code CoT   & 339K  & 2h   \\
            Scaling Code CoT    & 5.3M  & 8h   \\
            Question Reversal   & 4.6M  & 5h   \\
            Answer Generation   & 4.6M  & 40h  \\
            \midrule
            \textbf{Total (for 1.3M valid data)} & -- & \textbf{55h} \\
            \bottomrule
        \end{tabular}
    }
    \caption{Computation time for each stage in generating the \textsc{Caco-1.3M} dataset on a single 8$\times$A100 machine.}
    \label{tab:cost_analysis}
\end{table}
To quantify the efficiency of our method, we report the full computational cost of generating the \texttt{\texttt{Caco}}-1.3M dataset (Table~\ref{tab:cost_analysis}). 
All experiments were conducted on a single machine equipped with $8\times$ NVIDIA A100 GPUs. 
The pipeline consists of four main stages: \emph{unifying Code CoTs} ($339$K samples, $2$h), \emph{scaling Code CoTs} ($5.3$M samples, $8$h), \emph{question reversal} ($4.6$M samples, $6$h), and \emph{answer generation} ($4.6$M samples, $38.5$h), totaling approximately $55$ hours to produce $1.3$M verified samples. 
Importantly, the entire process relies solely on \emph{open-source} models, avoiding the substantial cost of proprietary API usage.

From a cost breakdown perspective, the majority of the runtime is consumed by the answer generation stage, which is \emph{unavoidable} in any instruction tuning or self-improvement setup. 
For example, prior works such as DartMath~\cite{dartmath} also incur comparable or higher costs in solution generation, particularly when sampling multiple candidate answers per prompt. 
The additional steps specific to \texttt{\texttt{Caco}}—Code CoT generation and question reversal—are lightweight (combined $\sim$16h), as natural language solutions are substantially longer than questions or Code CoTs. 

Overall, these results demonstrate that \texttt{\texttt{Caco}} can generate over one million verified, diverse reasoning samples in under three days on a single 8-GPU node, highlighting its strong \emph{scalability} and \emph{accessibility}. 
While we acknowledge that data-efficient methods have their merits, \texttt{\texttt{Caco}} is designed with a complementary focus: producing \emph{large-scale, diverse, and verifiable} reasoning data to support cross-domain generalization.

\section{Limitations}
\label{apx:limit}
Although \texttt{\texttt{Caco}} demonstrates strong capabilities in generating diverse reasoning paths and instructions, its performance is still limited by the predefined problem types used during training. The system may struggle when faced with highly innovative or unconventional problems, particularly those that do not align with the templates or problem categories used during training. As a result, generating high-quality code-based CoTs for more complex or uncommon problem types remains a challenge, potentially leading to biases in the distribution of generated data.

Additionally, while the code can be executed accurately, converting it back into human-readable natural language instructions may result in the loss of some details or require simplification, causing the final output to be less rich or specific than the original reasoning steps.

Furthermore, the generated code is primarily used for filtering data and not for final training purposes. It helps ensure the correctness and consistency of the reasoning process, but does not directly contribute to the final training dataset. In future work, it will be essential to explore how the generated code can be used to further improve the quality of the data and enhance the training process.

Currently, \texttt{\texttt{Caco}}'s application scope is focused mainly on mathematical and algorithmic reasoning tasks. Future work will need to explore extending it to broader domains, such as logical puzzles or STEM problem solving, which will require further effort.

\section{Future Works}
We outline three complementary directions: increasing \emph{difficulty}, expanding \emph{diversity}, and leveraging \texttt{\texttt{Caco}} in \emph{reinforcement learning (RL)}.

\paragraph{Raising Difficulty.}
Since the completion of this work, several math corpora with higher difficulty and quality than \textsc{DeepScaleR} and \textsc{BigMath} have emerged, such as \textsc{AM-Thinking-v1-Distilled} and the \textsc{DAPO} dataset. 
Starting from harder, cleaner seed sets is likely to further amplify the benefits of code-based augmentation. 
Concretely, we plan to (i) replace/augment the seed pool with high-difficulty problems (e.g., Olympiad-style, exam-grade items) and (ii) adopt hardness-aware sampling and adversarial program mutations during Code CoT generation. 

\paragraph{Expanding Diversity.}
As demonstrated in Section~\ref{ana:general}, our method generalizes beyond mathematics and applies naturally to domains with logical, symbolic, or procedural structure. 
While we discussed science and logic reasoning, a broader coverage (e.g., data reasoning, procedural planning, code debugging, diagram/physics problems, proofs) should allow \texttt{CodeGen} to learn richer templates and compose more diverse problems.
Furthermore, extending the framework beyond Python to support formal languages (e.g., Lean, Coq, or Wolfram Language) could enhance rigor and verifiability. 
We will (i) train multi-domain \texttt{CodeGen} with domain tags, (ii) design compositional templates that factor shared subroutines across domains.

\paragraph{Applications: Reinforcement Learning with Verifiable Rewards (RLVR).}
Recent RL-based training has shown strong gains for reasoning models but often depends critically on the correctness of reference answers. 
\texttt{\texttt{Caco}}’s executable traces provide a natural, low-noise reward signal. 
We will integrate \texttt{\texttt{Caco}} with RL by (i) deriving rewards from execution correctness (ii) employing a curriculum over program length and control-flow complexity. 
This combination targets scalable, verifiable RL training without heavy reliance on noisy external references.

\section{More Cases}
\label{apx:cases}
This section presents samples from the \method{} dataset, including the subsequence counting problem (Case \ref{case:subseq}), geometric sequences problem (Case \ref{case:geoseq}), permutation and combination problem (Case \ref{case:combine}), mathematical expression calculation problem (Case \ref{case:cal}), and analytical geometry problem (Case \ref{pmt:solidgeo}).

\setcounter{table}{0}
\renewcommand{\tablename}{Prompt}
\begin{table*}[h]
\begin{tcolorbox}[colback=gray!5,colframe=black!75, width=\textwidth, title=Code Generation Prompt]
\small
Given the following math problem in natural language, provide the complete code solution that solves the problem.

\vspace{1mm}
Requirements:
\begin{itemize}
    \item The final output of the program **must be the correct numerical or symbolic answer** to the problem.
    \item You must actually **compute** the result using Python code (e.g., using arithmetic or libraries like `sympy`), **not just explain in text or comments**.
    \item The code must define an `input` dictionary, call a function using that input, assign the result to a variable `output`, and finally `print(output)`.
    \item Please provide a complete, standalone executable script.
\end{itemize}
\vspace{1mm}

\#\#\# Example Math Problem:
\vspace{1mm}

A snail is at the bottom of a 20-foot well. Each day, it climbs up 3 feet, but at night, it slips back 2 feet. How many days will it take for the snail to reach the top of the well?
\vspace{1mm}

\#\#\# Example Code Solution:
\begin{lstlisting}[style=customPython]
def days_to_reach_top(well_height, climb_distance, slip_distance):
    days = 0
    current_height = 0

    while current_height < well_height:
        current_height += climb_distance
        if current_height >= well_height:
            break
        current_height -= slip_distance
        days += 1

    return days + 1

# Represent the input as a dictionary named 'input'
input = {{"well_height": 20, "climb_distance": 3, "slip_distance": 2}}
# Call the function with the input dictionary, assign the result to 'output'
output = days_to_reach_top(**input)
# Print the output
print(output)
\end{lstlisting}

Now, please provide the code solution for the following math problem directly. Make sure your code solution defines the input as a dictionary named input, calls the solution function using this dictionary, stores the result in a variable named output, and prints output.
\vspace{1mm}

\#\#\# Math Problem:

\{problem\}
\vspace{1mm}

\#\#\# Solution (Optional):

\{solution\}
\vspace{1mm}

\#\#\# Code Solution:
\end{tcolorbox}
\caption{Code Generation Prompt for solving a math problem using Python code.}
\label{pmt:math2code}
\end{table*} 
\begin{table*}[h]
\begin{tcolorbox}[colback=gray!5,colframe=black!75, width=\textwidth, title=Code Unifying Prompt]
\small
Given the following code and test function, please refactor the solution into the required format:

\#\#\# Example Output Format:
\begin{lstlisting}[style=customPython]
def add(a, b):
    return a + b

# Represent the input as a dictionary named 'input'
input = {{"a": 3, "b": 5}}
# Call the function with the input dictionary, assign the result to 'output'
output = add(**input)
# Print the output
print(output)
\end{lstlisting}
<answer>8</answer>

Code:

\{code\}

Test Function:

\{test\_code\}

Please refactor the code to follow the required format. 
\begin{itemize}
    \item The code must define an `input' dictionary, call a function using that input, assign the result to a variable `output', and finally `print(output)'.
    \item If there are multiple test cases in test function, just select one of them.
    \item Please provide a complete, standalone executable script.
\end{itemize}

\#\#\# Output:
\end{tcolorbox}
\caption{Prompt for refactoring code into the required input-output format.}
\label{pmt:unify_code}
\end{table*} 
\begin{table*}[h]
\begin{tcolorbox}[colback=gray!5,colframe=black!75, width=\textwidth, title=CodeGen Training Prompt]
\small
<|im\_start|>system

You are a helpful assistant.<|im\_end|>

<|im\_start|>user

\{code\}<|im\_end|>

\end{tcolorbox}
\caption{Prompt for training the CodeGen model.}
\label{pmt:codegen_train}
\end{table*} 

\begin{table*}[h]
\begin{tcolorbox}[colback=gray!5,colframe=black!75, width=\textwidth, title=CodeGen Sampling Prompt]
\small
<|im\_start|>system

You are a helpful assistant.<|im\_end|>

<|im\_start|>user

\end{tcolorbox}
\caption{Prompt for sampling from the trained CodeGen model.}
\label{pmt:codegen_sample}
\end{table*} 
\begin{table*}[h]
\begin{tcolorbox}[colback=gray!5,colframe=black!75, width=\textwidth, title=Question Back-translation Prompt]
\small
The code represents a solution to a math problem, and your task is to generate the original math problem that corresponds to the code.
\vspace{1mm}

\#\#\# Example Code:
\begin{lstlisting}[style=customPython]
def change_ref(amt, coins):
    if amt <= 0: return 0
    if amt != 0 and not coins: return float("inf")
    elif coins[0] > amt:
        return change_ref(amt, coins[1:])
    else:
        use_it = 1 + change_ref(amt - coins[0], coins)
        lose_it = change_ref(amt, coins[1:])
        return min(use_it, lose_it)

# Represent the input as a dictionary named 'input'
input = {"amt": 13, "coins": [1, 3, 5, 7]}
# Call the function with the input dictionary, assign the result to 'output'
output = change_ref(**input)
# Print the output
print(output)
\end{lstlisting}
\vspace{1mm}

\#\#\# Example Math Problem:

\vspace{1mm}
What is the minimum number of coins needed to make a total of 13 units using the available coin denominations of 1, 3, 5, and 7 units, each in unlimited supply?
\#\#\# End Problem
\vspace{1mm}

Please generate **Math Problem** based on the following code.
Ensure the generated problem is fully self-contained, solvable, and doesn't miss any necessary conditions or context.

\vspace{1mm}
You may add a concrete scenario or express the problem in different styles for diversity.  
\vspace{1mm}

\#\#\# Code:
\begin{verbatim}
{code}
\end{verbatim}

\#\#\# Math Problem:
\end{tcolorbox}
\caption{Question Back-translation Prompt. The prompt for generating a math problem based on a given code solution, where the generated problem should fully capture the conditions and context of the code.}
\label{pmt:code2math_q}
\end{table*} 

\begin{table*}[h]
\begin{tcolorbox}[colback=gray!5,colframe=black!75, width=\textwidth, title=Answer Generation Prompt]
\small
\#\#\# Instruction:

\{problem\}. Please reason step by step, and put your final answer within $\backslash$boxed\{\}.

\#\#\# Response:
\end{tcolorbox}
\caption{Instructions for generating step-by-step reasoning and the final answer enclosed in a boxed format.}
\label{pmt:code2math_a}
\end{table*} 
\begin{table*}[h]
\begin{tcolorbox}[colback=gray!5,colframe=black!75, width=\textwidth, title=Consistency Checking Prompt]
\small
Solution:\\
\{solution\}

Code:\\
\{code\}

Please determine if the logic of the code and the chain-of-thought in the solution are consistent.

Answer with a single word: "Yes" or "No".

Answer:

\end{tcolorbox}
\caption{Prompt for checking the consistency between the logic of the code and the chain-of-thought in the solution, where the answer is expected to be either "Yes" or "No".}
\label{pmt:consist}
\end{table*} 

\begin{table*}[h]
\begin{tcolorbox}[colback=gray!5,colframe=black!75, width=\textwidth, title=Solvability Checking Prompt]
\small
Problem: \\
\{problem\}

Please determine if the problem is solvable.

Answer with a single word: "Yes" or "No".

Answer:
\end{tcolorbox}
\caption{Prompt for determining the solvability of a given problem, where the answer is expected to be either "Yes" or "No".}
\label{pmt:solvable}
\end{table*}

\begin{table*}[h]
\begin{tcolorbox}[colback=gray!5,colframe=black!75, width=\textwidth, title=Correctness Checking Prompt]
\small
Problem: \\
\{problem\}

Solution:\\
\{solution\}

Please determine if the chain-of-thought of solution is correct.

Answer with a single word: "Yes" or "No".

Answer:
\end{tcolorbox}
\caption{Prompt for verifying the correctness of the chain-of-thought in a given solution, where the answer is expected to be either "Yes" or "No".}
\label{pmt:correct}
\end{table*} 

\begin{table*}[h]
\begin{tcolorbox}[colback=gray!5,colframe=black!75, width=\textwidth, title=Training Prompt]
\small
\#\#\# Instruction: \\
\{question\}
\vspace{2mm}

\#\#\# Response: \\
\{solution\}

\end{tcolorbox}
\caption{Training prompt of Caco model and baseline.}
\label{pmt:train_prompt}
\end{table*} 

\begin{table*}[h]
\begin{tcolorbox}[colback=gray!5,colframe=black!75, width=\textwidth, title=Evaluation Prompt]
\small
\#\#\# Instruction: \\
\{question\}
\vspace{2mm}

\#\#\# Response: Let's think step by step.

\end{tcolorbox}
\caption{Evaluation prompt of Caco model and baseline with step-by-step reasoning.}
\label{pmt:eval_prompt}
\end{table*} 
\setcounter{table}{0}
\renewcommand{\tablename}{Case}
\begin{table*}[h]
\begin{tcolorbox}[colback=gray!5,colframe=black!75, width=\textwidth, title=Subsequence Counting Problem]
\footnotesize
\textbf{Problem:} \\
How many unique subsequences can be formed from the list of numbers [1, 2, 3], where a subsequence is defined as any sequence that can be derived by deleting some or no elements from the list without changing the order of the remaining elements?

\vspace{1em}

\textbf{Code:}
\begin{lstlisting}[style=customPython]
def days_to_reach_top(well_height, climb_distance, slip_distance):
    days = 0
    current_height = 0

    while current_height < well_height:
        current_height += climb_distance
        if current_height >= well_height:
            break
        current_height -= slip_distance
        days += 1

    return days + 1

# Represent the input as a dictionary named 'input'
input = {{"well_height": 20, "climb_distance": 3, "slip_distance": 2}}
# Call the function with the input dictionary, assign the result to 'output'
output = days_to_reach_top(**input)
# Print the output
print(output)
\end{lstlisting}

\vspace{1em}

\textbf{Solution:} \\
We are asked to find the number of **unique subsequences** that can be formed from the list of numbers:

$$
[1, 2, 3]
$$

\#\#\# Step 1: Understand what a subsequence is

A **subsequence** is any sequence that can be derived by **deleting some or no elements** from the original list, **without changing the order** of the remaining elements.

For example, from [1, 2, 3], valid subsequences include:
[1],
[2],
[3],
[1, 2],
[1, 3],
[2, 3],
[1, 2, 3]

Also the empty subsequence (which is often considered, depending on the problem's context)

\#\#\# Step 2: Count all possible subsequences

Each element in the list can either be **included** or **excluded** in a subsequence. So for a list of length \( n \), there are \( 2^n \) total possible subsequences (including the empty subsequence).

Here, the list has 3 elements, so the total number of subsequences is:

$$
2^3 = 8
$$

\#\#\# Step 3: List all the subsequences to verify

Let’s list all 8 possible subsequences of [1, 2, 3]:

[] (empty subsequence),
[1],
[2],
[3],
[1, 2],
[1, 3],
[2, 3],
[1, 2, 3]

These are all the possible subsequences. All of them are **unique** because the elements are all distinct, and the order is preserved.

\#\#\# Final Answer:

$$
\boxed{8}
$$

There are **8 unique subsequences** that can be formed from the list [1, 2, 3].

\end{tcolorbox}
\caption{Subsequence Counting Problem: Counting the number of unique subsequences formed from a list of numbers.}
\label{case:subseq}
\end{table*}
\begin{table*}[h]
\begin{tcolorbox}[colback=gray!5,colframe=black!75, width=\textwidth, title=Geometric Sequence Problem]
\scriptsize
\textbf{Problem:} \\
If the real numbers $-1$, $a$, $b$, $c$, $-9$ form a geometric sequence, then $b=$ \_\_\_\_\_\_.

\textbf{Code:} 
\begin{lstlisting}[style=customPython]
from sympy import symbols, solve

def find_b(a, c):
    b = symbols('b')
    # Since a, b, c form a geometric sequence, b^2 = ac
    equation = b**2 - a*c
    solutions = solve(equation, b)
    # We need to determine the correct sign of b
    if solutions[0] * solutions[1] < 0:
        # If the product is negative, one of the solutions is negative
        # and the other is positive. We choose the one that is consistent
        # with the sequence -1, a, b, c, -9 being decreasing.
        if solutions[0] < 0:
            return solutions[0]
        else:
            return solutions[1]
    else:
        return solutions[0]

input = {"a": -1, "c": -9}
output = find_b(**input)
print(output)
\end{lstlisting}

\textbf{Solution:} \\
We are given that the real numbers $-1$, $a$, $b$, $c$, $-9$ form a **geometric sequence**. We are to find the value of $b$.

---

\#\#\# Step 1: Understand the structure of a geometric sequence

In a geometric sequence, each term is obtained by multiplying the previous term by a constant ratio $r$.

So, if the sequence is:

$$
-1,\ a,\ b,\ c,\ -9
$$

Then:

- $a = -1 \cdot r$

- $b = a \cdot r = (-1 \cdot r) \cdot r = -r^2$

- $c = b \cdot r = -r^2 \cdot r = -r^3$

- $-9 = c \cdot r = -r^3 \cdot r = -r^4$

---

\#\#\# Step 2: Use the last term to solve for $r$

We are told the last term is $-9$, so:

$$
-r^4 = -9
$$

Divide both sides by $-1$:

$$
r^4 = 9
$$

Now solve for $r$:

$$
r = \pm \sqrt[4]{9}
$$

Note that $\sqrt[4]{9} = \sqrt{\sqrt{9}} = \sqrt{3}$, so:

$$
r = \pm \sqrt{3}
$$

---

\#\#\# Step 3: Find $b = -r^2$

Recall that:

$$
b = -r^2
$$

Now compute $r^2$ for both values of $r$:

- If $r = \sqrt{3}$, then $r^2 = (\sqrt{3})^2 = 3$ \\
- If $r = -\sqrt{3}$, then $r^2 = (-\sqrt{3})^2 = 3$

So in both cases:

$$
b = -r^2 = -3
$$

---

\#\#\# Final Answer:

$$
\boxed{-3}
$$

\end{tcolorbox}
\caption{Geometric Sequence Problem: Solving for $b$ in a geometric sequence with given terms.}
\label{case:geoseq}
\end{table*}

\begin{table*}[h]
\begin{tcolorbox}[colback=gray!5,colframe=black!75, width=\textwidth, title=Counting and Combinatorics]
\footnotesize
\textbf{Problem:} \\
Using the digits 1 and 3 to form a four-digit number, and both digits 1 and 3 must appear at least once, there are a total of $\_\_\_\_\_\_\_$ such four-digit numbers.

\textbf{Code:} \\
\begin{lstlisting}[style=customPython]
from itertools import product

def count_four_digit_numbers():
    count = 0
    # Generate all 4-digit combinations of 1 and 3
    for combo in product([1, 3], repeat=4):
        # Ensure both 1 and 3 appear at least once
        if 1 in combo and 3 in combo:
            count += 1
    return count

# Represent the input as a dictionary named 'input'
input = {}
# Call the function with the input dictionary, assign the result to 'output'
output = count_four_digit_numbers(**input)
# Print the output
print(output)
\end{lstlisting}

\textbf{Solution:} \\
We are asked to find how many **four-digit numbers** can be formed using only the digits **1** and **3**, such that **both digits 1 and 3 appear at least once**.

---

\#\#\# Step 1: Total number of four-digit numbers using digits 1 and 3

Each of the 4 positions in the number can be filled with either **1** or **3**, so:

$$
\text{Total combinations} = 2^4 = 16
$$

So, there are **16** total four-digit numbers that can be formed using only digits 1 and 3.

---

\#\#\# Step 2: Subtract the cases where only one digit is used

We need to exclude the numbers that **do not contain both digits**. That is, the cases where:

- All digits are **1** → number: 1111
- All digits are **3** → number: 3333

So, there are **2** such numbers that we need to exclude.

---

\#\#\# Step 3: Compute the valid numbers

$$
\text{Valid numbers} = \text{Total combinations} - \text{Invalid combinations}
$$
$$
= 16 - 2 = \boxed{14}
$$

---

\#\#\# Final Answer:

$$
\boxed{14}
$$

There are **14** four-digit numbers that can be formed using the digits 1 and 3, with both digits appearing at least once.

\end{tcolorbox}
\caption{Counting and Combinatorics problem: Counting the number of valid four-digit numbers formed by the digits 1 and 3 where both digits appear at least once. The table includes the problem description, step-by-step solution, and corresponding Python code to calculate the number of valid combinations.}
\label{case:combine}
\end{table*}
\begin{table*}[h]
\begin{tcolorbox}[colback=gray!5,colframe=black!75, width=\textwidth, title=Mathematical Expression Calculation]
\footnotesize
\textbf{Problem:} \\
Calculate: 
\[
\sqrt{12} + |1 - \sqrt{3}| + (\pi - 2023)^{0}
\]

\textbf{Code:} \\
\begin{lstlisting}[style=customPython]
import math

def calculate_expression():
    sqrt_12 = math.sqrt(12)
    abs_value = abs(1 - math.sqrt(3))
    power_0 = (math.pi - 2023) ** 0
    result = sqrt_12 + abs_value + power_0
    return result

# Represent the input as a dictionary named 'input'
input = {}
# Call the function with the input dictionary, assign the result to 'output'
output = calculate_expression(**input)
# Print the output
print(output)
\end{lstlisting}

\textbf{Solution:} \\
We are asked to calculate the expression:

$$
\sqrt{12} + |1 - \sqrt{3}| + (\pi - 2023)^0
$$

Let’s break it down step by step.

---

\#\#\# Step 1: Simplify $ \sqrt{12} $

We can simplify the square root:

$$
\sqrt{12} = \sqrt{4 \cdot 3} = \sqrt{4} \cdot \sqrt{3} = 2\sqrt{3}
$$

---

\#\#\# Step 2: Simplify $ |1 - \sqrt{3}| $

We know that $ \sqrt{3} \approx 1.732 $, so:

$$
1 - \sqrt{3} \approx 1 - 1.732 = -0.732
$$

Since the result is negative, the absolute value is:

$$
|1 - \sqrt{3}| = \sqrt{3} - 1
$$

---

\#\#\# Step 3: Simplify $ (\pi - 2023)^0 $

Any non-zero number raised to the power of 0 is 1. Since $ \pi \approx 3.1416 $, and $ \pi - 2023 \neq 0 $, we have:

$$
(\pi - 2023)^0 = 1
$$

---

\#\#\# Step 4: Add all the simplified parts

Now, add the three parts together:

$$
\sqrt{12} + |1 - \sqrt{3}| + (\pi - 2023)^0 = 2\sqrt{3} + (\sqrt{3} - 1) + 1
$$

Simplify:

$$
2\sqrt{3} + \sqrt{3} - 1 + 1 = 3\sqrt{3}
$$

---

\#\#\# Final Answer:

$$
\boxed{3\sqrt{3}}
$$

\end{tcolorbox}
\caption{Mathematical Expression Calculation: Solving the expression involving square roots, absolute values, and exponentiation. The table provides a detailed breakdown of the problem-solving steps and the corresponding Python code implementation.}
\label{case:cal}
\end{table*}

\vspace{-8mm}
\begin{table*}[h]
\vspace{-8mm}
\begin{tcolorbox}[colback=gray!5,colframe=black!75, width=\textwidth, title=Tangency Condition of Line and Parabola]
\scriptsize
\textbf{Problem:} \\
Given the line $x - y - 1 = 0$ and the parabola $y = ax^{2}$ are tangent to each other, find the value of $a$.

\textbf{Code:} 
\vspace{-0.5mm}
\begin{lstlisting}[style=customPython]
import sympy as sp

def find_tangent_coefficient():
    x, y, a = sp.symbols('x y a')
    line_eq = x - y - 1
    parabola_eq = y - a*x**2
    # Substitute y from the line equation into the parabola equation
    substituted_eq = parabola_eq.subs(y, x - 1)
    # Simplify the equation
    simplified_eq = sp.simplify(substituted_eq)
    # For the line and parabola to be tangent, the quadratic equation must have exactly one solution.
    discriminant = sp.discriminant(simplified_eq, x)
    # Solve for a
    a_value = sp.solve(discriminant, a)
    return a_value[0]

# Represent the input as a dictionary named 'input'
input = {}
# Call the function with the input dictionary, assign the result to 'output'
output = find_tangent_coefficient(**input)
# Print the output
print(output)
\end{lstlisting}

\textbf{Solution:} \\
We are given:

- A **line**: $ x - y - 1 = 0 $

- A **parabola**: $ y = ax^2 $

We are told that the **line and the parabola are tangent to each other**, which means they intersect at **exactly one point** (i.e., the system has exactly one solution).

---

\#\#\# Step 1: Solve the system of equations

We are given:

1. $ x - y - 1 = 0 \Rightarrow y = x - 1 $

2. $ y = ax^2 $

Substitute the expression for $ y $ from the line into the parabola:

$$
x - 1 = ax^2
$$

Rewriting:

$$
ax^2 - x + 1 = 0
$$

This is a **quadratic equation** in $ x $:

$$
ax^2 - x + 1 = 0
$$

---

\#\#\# Step 2: Use the condition for tangency

For the line and the parabola to be **tangent**, the quadratic equation must have **exactly one solution**. That means the **discriminant** of the quadratic must be zero.

Recall that for a quadratic $ ax^2 + bx + c = 0 $, the discriminant is:

$$
\Delta = b^2 - 4ac
$$

In our case, the quadratic is:

$$
ax^2 - x + 1 = 0
$$

So:

$ a = a $, $ b = -1 $, $ c = 1 $

Now compute the discriminant:

$$
\Delta = (-1)^2 - 4(a)(1) = 1 - 4a
$$

Set the discriminant equal to zero (for tangency):

$$
1 - 4a = 0
$$

Solve for $ a $:

$$
4a = 1 \Rightarrow a = \frac{1}{4}
$$

\#\#\# Final Answer:

$$
\boxed{\frac{1}{4}}
$$

\end{tcolorbox}
\caption{Tangency Condition of Line and Parabola: Finding the value of $a$ for the line and parabola to be tangent.}
\label{pmt:solidgeo}
\end{table*}

\clearpage

\end{document}